\documentclass{article}
\usepackage{ijcai25}

\usepackage{times}
\usepackage{soul}
\usepackage{url}
\usepackage[hidelinks]{hyperref}
\usepackage[utf8]{inputenc}
\usepackage[small]{caption}
\usepackage{subfigure}
\usepackage{graphicx}
\usepackage{mathtools}
\usepackage{amsfonts}
\usepackage{amsthm}
\usepackage{amsmath, amssymb, mathrsfs}
\usepackage{booktabs}
\usepackage{algorithm}
\usepackage{algorithmic}
\usepackage[switch]{lineno}
\usepackage{natbib}

\newtheorem{theorem}{Theorem}

\title{Constructing Non-Markovian Decision Process via History Aggregator}

\author{
Yongyi Wang$^{1,2}$
\and
Wenxin Li$^{1,2}$\\
\affiliations
$^1$AILab, School of Computer Science, Peking University, Beijing 100871, China.\\
$^2$National Key Laboratory for Multimedia Information Processing, School of Computer Science, Peking University, Beijing 100871, China.\\
\emails
\{wangyongyi, lwx\}@pku.edu.cn
}

\begin{document}

\maketitle

\begin{abstract}
In the domain of algorithmic decision-making, non-Markovian dynamics manifest as a significant impediment, especially for paradigms such as Reinforcement Learning (RL), thereby exerting far-reaching consequences on the advancement and effectiveness of the associated systems.
Nevertheless, the existing benchmarks are deficient in comprehensively assessing the capacity of decision algorithms to handle non-Markovian dynamics.
To address this deficiency, we have devised a generalized methodology grounded in category theory. Notably, we established the category of Markov Decision Processes (MDP) and the category of non-Markovian Decision Processes (NMDP), and proved the equivalence relationship between them. This theoretical foundation provides a novel perspective for understanding and addressing non-Markovian dynamics.
We further introduced non-Markovianity into decision-making problem settings via the History Aggregator for State (HAS). With HAS, we can precisely control the state dependency structure of decision-making problems in the time series.
Our analysis demonstrates the effectiveness of our method in representing a broad range of non-Markovian dynamics. This approach facilitates a more rigorous and flexible evaluation of decision algorithms by testing them in problem settings where non-Markovian dynamics are explicitly constructed.
\end{abstract}

\section{Introduction}
\label{sec:intro}
Decision-making algorithms, such as RL, generally demand that the problem satisfy the Markov assumption, which posits that the current observation encapsulates all accessible information for decision-making, rendering past observations irrelevant.
However, many real-world systems do not adhere to this assumption. For example, control systems often exhibit non-Markovianity due to limited sensor information about internal states~\citep{whitehead1995reinforcement}.
Similarly, open quantum systems display non-Markovian characteristics due to complex interactions with their external environment~\citep{RevModPhys.89.015001}.
Such non-Markovian environments are prevalent in real-world applications, including human physiology, biological systems, and material science~\citep{gupta2021non}.

For the characterization and resolution of non-Markovian decision problems, in addition to NMDP, several theoretical models have emerged, including linear-dynamic-logic-based Regular Decision Process (RDP)~\citep{brafman2019regular,abadi2020learning}, automaton-based Reward Machine (RM)~\citep{camacho2019ltl,rens2020learning}, and Markov Abstraction~\citep{hutter2009feature,maillard2011selecting,veness2011monte,nguyen2013competing,lattimore2013sample,majeed2018q,ronca2022markov}, among others.
These models are utilized either to model the state transition dynamics or the reward mechanisms of non-Markovianity, but the way they model non-Markovianity varies, thereby causing inconvenience to the research of decision algorithms.
In response, we establish the MDP category and the NMDP category, providing a unified and rigorous view within the framework of category theory.
Under this framework, any general approach that converts NMDPs into MDPs while preserving the transition dynamics can be seen as a functor from the NMDP category to the MDP category.
This framework not only encapsulates the essence of preceding research but also paves the way for devising novel strategies to tackle non-Markovian decision problems.
Moreover, the utilization of the category theory perspective facilitates a deeper understanding of the relationship between MDP and NMDP.
We prove that the MDP category and the NMDP category are equivalent categories based on the intuition that any NMDP can be transformed into an MDP, given that the entire history can be regarded as a state.

Based on the theoretical models mentioned above, numerous algorithms have emerged to solve non-Markovian decision problems. However, an algorithm that is both commonly accepted and highly effective for NMDPs remains elusive.
The evaluation of such algorithms relies on non-Markovian decision problems, which poses a set of challenges.
For instance, \cite{gaon2020reinforcement} utilizes a multi-armed bandit and robot grid world with non-Markovian rewards; \cite{gupta2021non} employs a blood glucose control simulation environment; \cite{ronca2022markov} uses Rotating MAB, Malfunction MAB, Enemy Corridor, Reset - Rotating MAB, and Flickering grid; \cite{dohmen2022inferring} uses an office grid-world scenario; \cite{qin2024learning} uses modified Mujoco environments; \cite{chandak2024reinforcement} uses modified Cartpole and mountain Car, along with a controlled non-Markovian random walk.
The non-Markovian environments described in the literature, which are scattered and unsystematic, originate from diverse fields.
Moreover, most of these environments are either obtained through simple construction or derived from masking some part of the observation in existing MDP, that is, by transforming it into a Partially Observable MDP (POMDP), which may change the internal difficulty because of information loss in the observation.
Their simplicity and lack of standardization impede the development of a universal benchmark, thereby limiting the ability to comprehensively evaluate how effectively algorithms can manage non-Markovian dynamics.

To address the issue, we have developed a method for constructing NMDPs from MDPs.
We use the History Aggregator for State (HAS) and History Aggregator for Reward (HAR) to introduce non-Markovianity into MDPs' state transition dynamics and reward mechanisms.
Given that transition dynamics is more essential in non-Markovianity, this paper focuses on HAS.
When HAS meets the reversibility condition, the original MDP's essence is preserved during transformation.
Specifically, this condition enables the original state to be decoded from the NMDP's complete history without information loss.
Using reversible HAS allows the constructed NMDP to more precisely evaluate decision-making algorithms' ability to remember and decode history to obtain the original state.
Since handling complex temporal dependencies through memorization and decoding is crucial for addressing non-Markovianity, our method, using different HAS, constructs diverse temporal dependency structures, providing greater flexibility and applicability.

We propose two special kinds of easily-implementable reversible HAS. One aggregates history by applying a binary group operator between states, and the other by using auxiliary sequences and ring operators.
The idea behind the first one is that directly adding vector states raises the order of Markov dependence by one.
For the second, its intuition is derived from the convolution operation in signal processing.
We use an auxiliary sequence to assign weights to states and then aggregate them.
In actual problem scenarios, states are generally representable as real vectors.
Thus, these two types of HAS can be straightforwardly implemented using linear algebra operators.
Moreover, the non-Markovian environments constructed by these two methods share the same interface as the original Markovian environments.
This characteristic enables the direct application of the same decision-making algorithm to solve problems in both types of environments without modification.
Consequently, it becomes possible to conduct a pure assessment of the performance degradation of decision-making algorithms resulting from the introduction of non-Markovianity.
There is no need to consider changes in the inherent complexity of the decision-making problem itself.
This is because the construction based on the HAS only imposes requirements on the decision-making algorithm's capacity for history memorization and decoding.

Building on the challenges associated with constructing non-Markovian decision problems as discussed above, the subsequent chapters of our work are structured to provide solutions.
In Chapter \ref{sec:prelim}, we introduce the fundamentals of category theory, MDP, and NMDP, along with the basics of algebra. This serves as essential background for understanding our work.
In Chapter \ref{sec:category}, we formally describe and prove the equivalence of the MDP and NMDP categories. This not only unifies the two concepts but also naturally leads to the method of constructing NMDPs in the subsequent chapter.
Chapter \ref{sec:method} presents the definition of HAS, two construction methods using reversible HAS, and an analysis of their properties. These methods innovatively introduce non-Markovianity while preserving the original nature of the problem.
Although the main contribution of our work is theoretical,  we convert some classical MDPs into NMDPs and conduct several empirical experiments using some RL algorithms in Chapter \ref{sec:exp}.
This validates our theoretical framework and demonstrates its applicability.
The Appendix provides mathematical proofs, illustrative examples, intuitive explanations of key concepts, and links to the code. This allows the main text to focus on the key ideas without being cluttered with details.
Overall, our work offers a comprehensive, novel, and practical solution for constructing non-Markovian decision problems.


\newcommand{\funcat}{\mathbf{Fun}}
\newcommand{\mton}{\funcat(\mdpcat,\nmdpcat)}
\newcommand{\ntom}{\funcat(\nmdpcat,\mdpcat)}
\newcommand{\nton}{\funcat(\nmdpcat,\nmdpcat)}
\newcommand{\nat}{\mathbb{N}}
\newcommand{\natplus}{\nat^{+}}
\newcommand{\real}{\mathbb{R}}
\newcommand{\mdp}{\mathcal{M}}
\newcommand{\nmdp}{\mathcal{N}}
\newcommand{\mdpcat}{\mathfrak{M}}
\newcommand{\nmdpcat}{\mathfrak{N}}
\newcommand{\mdpdef}{\langle\rho_0, S, A, \{T_t\}_{t=0}^\infty\rangle}
\newcommand{\mdpdefp}{\langle\rho_0^\prime, S^\prime, A^\prime, \{T^\prime_t\}_{t=0}^\infty\rangle}
\newcommand{\nmdpdef}{\langle\rho_0, S, A, \{T_t\}_{t=0}^\infty\rangle}
\newcommand{\nmdpdefp}{\langle\rho_0^\prime, S^\prime, A^\prime, \{T^\prime_t\}_{t=0}^\infty\rangle}
\newcommand{\morphism}[1]{(#1_{S},#1_{A},#1_{\real})}
\newcommand{\ha}[2]{\mathscr{A}_{#1}^{#2}}
\newcommand{\hadef}[2]{\{\mathscr{A}_{#1,t}^{#2}\}_{t=0}^\infty}
\newcommand{\hb}[2]{\mathscr{B}_{#1}^{#2}}
\newcommand{\hbdef}[2]{\{\mathscr{B}_{#1,t}^{#2}\}_{t=0}^\infty}
\newcommand{\laast}{\overset{\scriptscriptstyle\leftarrow}{\ast}}
\newcommand{\raast}{\overset{\scriptscriptstyle\rightarrow}{\ast}}
\newtheorem{definition}{Definition}
\newtheorem{corollary}{Corollary}

\section{Preliminaries}
\label{sec:prelim}
\subsection{Notations}

In the following, we denote by $\Delta_X\coloneqq\{p\mid p:X\to[0,\infty),~\sum_{x\in X}p(x)=1\}$
the set of all probability distributions on the set $X$, denote by $f^{\langle n\rangle}$ an $n$-variable function created from the unary function $f$ in the manner of $f^{\langle n\rangle}(x_1,x_2,\cdots,x_n)\coloneqq(f(x_1),f(x_2),\cdots,f(x_n))$, and denote by $x_{m:n}~(m,n\in\nat,m\le n)$ a finite sequence $\{x_i\}_{i=m}^n$. We do not distinguish between $x_{n:n}$ and $x_n$ in the following.

\subsection{Category Theory}
A category is a system of related objects. The objects do not live in isolation: there is some notion of morphism, or equivalently, map, between objects, binding them together~\citep{leinster2014basic}.
Category theory delves into the relationships between objects and morphisms.
Objects can be of any nature, and morphisms denote the relations or transformations linking these objects.
For example, all groups together with all group homomorphisms form a category of groups, and any partially ordered set can also form a category.
This theory offers a potent framework for apprehending mathematical structures.
By distilling common traits, category theory allows for a unified analysis across diverse mathematical domains.
It uncovers profound connections and streamlines intricate concepts, presenting a more coherent view of mathematics.
\begin{definition}[\textbf{Category}]
A category $\mathfrak{C}$ consists of:
\begin{itemize}
    \item a collection $ob(\mathfrak{C})$ of \textbf{objects}; 
    \item for each $\mathcal{A},\mathcal{B}\in ob(\mathfrak{C})$, a collection $\mathfrak{C}(\mathcal{A},\mathcal{B})$ of \textbf{morphisms} from $\mathcal{A}$ to $\mathcal{B}$; 
    \item for each $\mathcal{A},\mathcal{B},\mathcal{C}\in ob(\mathfrak{C})$, a function $$\begin{array}{ccl}\mathfrak{C}(\mathcal{A},\mathcal{B})\times\mathfrak{C}(\mathcal{B},\mathcal{C}) & \to & \mathfrak{C}(\mathcal{A},\mathcal{C}) \\ (g,f) & \mapsto & g\circ f\end{array}$$ called \textbf{composition};
    \item for each $\mathcal{A}\in ob(\mathfrak{C})$, an element $1_\mathcal{A}\in\mathfrak{C}(\mathcal{A},\mathcal{A})$, called the \textbf{identity} on $\mathcal{A}$,
\end{itemize}
satisfying the following axioms:
\begin{itemize}
    \item \textbf{associativity}: for each $f\in\mathfrak{C}(\mathcal{A},\mathcal{B})$, $g\in\mathfrak{C}(\mathcal{B},\mathcal{C})$, $h\in\mathfrak{C}(\mathcal{C},\mathcal{D})$ we have $(h\circ g)\circ f=h\circ(g\circ f)$;
    \item \textbf{identity}: for each $f\in\mathfrak{C}(\mathcal{A},\mathcal{B})$, we have $f=f\circ 1_\mathcal{A}=1_\mathcal{B}\circ f$.
\end{itemize}
\end{definition}
Just as morphisms convey intra-category object relations or transformations, functors represent inter-category ones.

\begin{definition}[\textbf{Functor}] Let $\mathfrak{A},\mathfrak{B}$ be categories. A \textbf{functor} $\mathbf{F}:\mathfrak{A}\to\mathfrak{B}$ consists of: 
\begin{itemize}
    \item a function $ob(\mathfrak{A})\to ob(\mathfrak{B})$ written as $\mathcal{A}\mapsto\mathbf{F}(\mathcal{A})$;
    \item for each $\mathcal{A},\mathcal{A}^\prime\in\mathfrak{A}$, a function $$\mathfrak{A}(\mathcal{A},\mathcal{A}^\prime)\to\mathfrak{B}(\mathbf{F}(\mathcal{A}),\mathbf{F}(\mathcal{A}^\prime))$$ written as $f\mapsto\mathbf{F}(f)$;
\end{itemize}
satisfying the following axioms:
\begin{itemize}
    \item $\mathbf{F}(f^\prime\circ f)=\mathbf{F}(f^\prime)\circ\mathbf{F}(f)$ if $\mathcal{A}\xrightarrow{f}\mathcal{A}^\prime\xrightarrow{f^\prime}\mathcal{A}^{\prime\prime}$ in $\mathfrak{A}$;
    \item $\mathbf{F}(1_\mathcal{A})=1_{\mathbf{F}(\mathcal{A})}$ if $\mathcal{A}\in\mathfrak{A}$.
\end{itemize}
\end{definition}
The collection of functors from $\mathfrak{A}$ to $\mathfrak{B}$ also forms a category $\funcat(\mathfrak{A},\mathfrak{B})$.

\subsection{Algebra}
\begin{definition}[\textbf{Semigroup}]
Let $S$ be a non-empty set, $\cdot: S\times S\to S$ is a binary operator, if the following axioms hold:
\begin{itemize}
    \item \textbf{closure}: $\forall a,b\in S,~a\cdot b\in S$;
    \item \textbf{associativity}: $\forall a,b,c\in S,~a\cdot(b\cdot c)=(a\cdot b)\cdot c$.
\end{itemize}
Then $(S,\cdot)$ forms a semigroup.
\end{definition}
If there exists a \textbf{unit} $e\in S$ in a semigroup $(S,\cdot)$ such that $\forall a\in S,~a\cdot e=e\cdot a=a$, then $(S,\cdot)$ becomes a \textbf{monoid}.
\begin{definition}[\textbf{Group}]
Let $(G,\cdot)$ be a monoid, $e$ is the unit, if the following axiom holds:
\begin{itemize}
    \item \textbf{inverse}: $\forall a\in G,~\exists a^{-1}\in G,~a^{-1}\cdot a=a\cdot a^{-1}=e$.
\end{itemize}
Then $(G,\cdot)$ forms a group.
\end{definition}
If the operator in a group $(G,\cdot)$ is commutative, i.e. $\forall a,b\in G,~a\cdot b=b\cdot a$, then $(G,\cdot)$ is an \textbf{abelian group}.

\begin{definition}[\textbf{Ring}]
Let $R$ be a non-empty set, if $(R,\cdot)$ forms a semigroup, $(R,+)$ forms an abelian group with unit $0\in R$, and the following axioms hold:
\begin{itemize}
    \item \textbf{left distributivity}: $\forall a,b,c\in R,~a\cdot(b+c)=a\cdot b+a\cdot c$;
    \item \textbf{right distributivity}: $\forall a,b,c\in R,~(b+c)\cdot a=b\cdot a+c\cdot a$.
\end{itemize}
Then $(R,+,\cdot)$ forms a ring.
\end{definition}
If the operator $\cdot$ in ring $(R,+,\cdot)$ is commutative, then $(R,+,\cdot)$ becomes a \textbf{commutative ring}, and if $(R,\cdot)$ is a monoid with unit $1$, then $(R,+,\cdot)$ becomes a \textbf{unital ring} with unit $1$.
\begin{definition}[\textbf{Left $R$-Module}]
Let $(R,+,\cdot)$ be a ring, $(M,\oplus)$ be an abelian group, if there exists a map $R\times M\to M$ under which the image of any $(r,m)\in R\times M$ is denoted as $rm\in M$, and the following axioms hold:
\begin{itemize}
\item \textbf{$R$-distributivity}: $$\forall r\in R,~\forall m,m^\prime\in M,~r(m\oplus m^\prime)=rm\oplus rm^\prime$$
\item \textbf{$M$-distributivity}: $$\forall r,r^\prime\in R,~\forall m\in M,~(r+r^\prime)m=rm\oplus r^\prime m$$
\item \textbf{associativity}: $$\forall r,r^\prime\in R,~\forall m\in M,~r(r^\prime m)=(r\cdot r^\prime)m$$
\item \textbf{identity}: if $(R,+,\cdot)$ is a unital ring with unit $1$, then $\forall m\in M, ~1m=m$.
\end{itemize}
Then $M$ is called a left $R$-module.
\end{definition}
A left $R$-module is a generalization of a linear space. When $R$ becomes a field, left $R$-module $M$ becomes a linear space.

\subsection{MDP and NMDP}
Before setting up the MDP and NMDP categories, we pin down their definitions.
\begin{definition}[\textbf{Markov Decision Process (MDP)}]
An MDP $\mdp$ is a tuple $\mdpdef$ consisting of:
    \begin{itemize}
        \item Initial state distribution $\rho_0\in\Delta_S$;
        \item State set $S$;
        \item Action set $A$;
        \item Transition dynamics at time $t$:
        $$T_t:S\times A\to\Delta_{S\times\real}$$
    \end{itemize}
\end{definition}
A \textbf{policy} for an MDP at time $t$ is defined as $\pi_t:S\to\Delta_A$.
An MDP is called \textbf{degenerate} if $\exists s,s^\prime\in S, ~\forall t\in\nat, ~\forall a\in A, ~T_t(s,a)=T_t(s^\prime,a)$, which means there are indistinguishable states in $S$.

Note that, by definition, an MDP is not necessarily time-homogeneous.
However, any non-time-homogeneous MDP can be converted into a time-homogeneous MDP by extending the state set $S$ to be a subset of $\nat\times S$, where each state is associated with its timestep, i.e., $s_t\mapsto(t,s_t)$ and $T:((t,s_t),a)\mapsto T_t(s_t,a)$.
\begin{definition}[\textbf{Non-Markov Decision Process (NMDP)}]
An NMDP $\nmdp$ is a tuple $\nmdpdef$ consisting of:
    \begin{itemize}
        \item Initial state distribution $\rho_0\in\Delta_S$;
        \item State set $S$;
        \item Action set $A$;
        \item Transition dynamics at time $t$:
    $$T_t: S^{t+1}\times A^{t}\times\real^{t}\times A\to\Delta_{S\times\real}$$
    \end{itemize}
\end{definition}
A \textbf{policy} for an NMDP at time $t$ is $\pi_t:S^{t+1}\times A^{t}\times\real^{t}\to\Delta_A$.
The \textbf{history set} \textbf{at time $t$} for an MDP or an NMDP  is $H_t\coloneqq S^{t+1}\times A^{t}\times\real^{t}$, with which transition dynamics and policy of an NMDP can be rewritten as $T_t: H_t\times A\to\Delta_{S\times\real}$,~~$\pi_t: H_t\to\Delta_A$.
The \textbf{history set} $H$ for an MDP or an NMDP is defined as the disjoint union of all $H_t$, i.e., $H\coloneqq\bigsqcup_{t=0}^\infty H_t$.
The \textbf{proper prefix} relation between $h$ and $h^\prime$ is denoted as $h\prec h^\prime$, indicating that $h$ is a proper prefix of $h^\prime$, i.e. $h=(s_{0:t},a_{0:t-1},r_{0:t-1})$ and $h^\prime=(s_{0:t+k},a_{0:t+k-1},r_{0:t+k-1}), k\in\natplus$.

With the concept of the history set for both MDPs and NMDPs, we can define operators that extract states, actions, and rewards from a history.

\begin{definition}[\textbf{States, Actions, and Rewards Extraction Operators}]
The states, actions, and rewards extraction operators are:
\begin{itemize}
    \item $\mathscr{E}_S: h_t\mapsto s_{0:t}~~(\forall t\in\nat,~\forall h_t\in H_t)$;
    \item $\mathscr{E}_A: h_t\mapsto a_{0:t-1}~~(\forall t\in\natplus,~\forall h_{t}\in H_{t})$;
    \item $\mathscr{E}_{\real}: h_t\mapsto r_{0:t-1}~~(\forall t\in\natplus,~\forall h_{t}\in H_{t})$.
\end{itemize}
\end{definition}

\begin{definition}[\textbf{Latest State, Action, and Reward Extraction Operators}]
The latest state, action, and reward extraction operators at time $t$ are:
\begin{itemize}
    \item $\mathscr{L}_{S,t}: h_t\mapsto s_t~~(\forall t\in\nat,~\forall h_t\in H_t)$;
    \item $\mathscr{L}_{A,t}: h_{t+1}\mapsto a_{t}~~(\forall t\in\nat,~\forall h_{t+1}\in H_{t+1})$;
    \item $\mathscr{L}_{\real,t}: h_{t+1}\mapsto r_{t}~~(\forall t\in\nat,~\forall h_{t+1}\in H_{t+1})$.
\end{itemize}
\end{definition}

\section{Equivalence of MDP and NMDP Category}
\label{sec:category}
In this chapter, we present the definitions of the MDP category and the NMDP category, together with the theorem regarding their equivalence relationship. The relevant proofs are provided in the Appendix.

\begin{definition}[\textbf{MDP Category}]
 $\mdpcat$ is a category where:
    \begin{itemize}
        \item \textbf{Objects}: The objects are MDPs in the form of \newline $\mdp=\mdpdef$;
        \item \textbf{Morphisms}: A morphism $\phi=\morphism{\phi}$ from $\mdp=\mdpdef$ to $\mdp^\prime=\mdpdefp$ where $\phi_S: S\to S^\prime$, $\phi_A: A\to A^\prime$, $\phi_\real: \real\to\real$, satisfies the following properties:
        \begin{itemize}
            \item $\rho_0=\rho^\prime_0\circ\phi_S$;
            \item $T_t(s,a)=((T_t^\prime\circ(\phi_S,\phi_A))(s,a))\circ (\phi_S,\phi_\real)$,\newline
            $(\forall t\in\nat, \forall s\in S, \forall a\in A)$.
        \end{itemize}
        \textbf{Identity}: $1_\mdp=\morphism{1}$;\newline
        \textbf{Composition}: composition of $\morphism{\phi^\prime}$ and $\morphism{\phi}$ is $(\phi^\prime_S\circ\phi_S,\phi_A^\prime\circ\phi_A,\phi_\real^\prime\circ\phi_\real)$.
    \end{itemize}
\end{definition}

\begin{definition}[\textbf{NMDP Category}]
 $\nmdpcat$  is a category where:
    \begin{itemize}
        \item \textbf{Objects}: The objects are NMDPs in the form of \newline $\nmdp=\nmdpdef$;
        \item \textbf{Morphisms}: A morphism $\psi=\morphism{\psi}$ from $\nmdp=\nmdpdef$ to $\nmdp^\prime=\nmdpdefp$ where $\psi_S: S\to S^\prime$, $\psi_A: A\to A^\prime$, $\psi_\real: \real\to\real$, satisfies the following properties:
        \begin{itemize}
            \item $\rho_0=\rho_0^\prime\circ\psi_S$;
            \item $T_t(h_t,a)=((T_t^\prime\circ(\psi_{H_t},\psi_A))(h_t,a))\circ(\psi_S,\psi_\real)$,\newline
            $~(\forall t\in\natplus, \forall h_t\in H_t, \forall a\in A)$,\newline
            where $\psi_{H_t}=(\psi_S^{\langle t+1\rangle},\psi_A^{
        \langle t\rangle},\psi_\real^{\langle t\rangle})$.
        \end{itemize}
        \textbf{Identity}: $1_\nmdp=\morphism{1}$;\newline
        \textbf{Composition}: composition of $\morphism{\psi^\prime}$ and $\morphism{\psi}$ is $(\psi^\prime_S\circ\psi_S,\psi_A^\prime\circ\psi_A,\psi_\real^\prime\circ\psi_\real)$.
    \end{itemize}
\end{definition}

A morphism in either category represents that the transition dynamics of the source decision process can be "simulated" by the target.

As discussed in Chapter \ref{sec:intro}, previous algorithmic works have sought functors in $\ntom$, which can generally be summarized as approximations of the Markov abstraction functor $\mathbf{M}\in\ntom$, utilizing various techniques to compress history.
In contrast to what is described in the existing literature, our definition of Markov abstraction retains the entire history without any compression.

\begin{definition}[\textbf{Markov Abstraction}]
 Markov abstraction is a functor $\mathbf{M}:\nmdpcat\to\mdpcat$. For any $\nmdp=\nmdpdef\in\nmdpcat$ there exists $\mdp^\prime=\mathbf{M}(\nmdp)=\mdpdefp\in\mdpcat$, $\rho_0^\prime=\rho_0$, $A^\prime=A$, which satisfies the following properties:
    \begin{itemize}
        \item $S^\prime=H$, where $H$ is the history set of $\nmdp$;
        \item $T_t^\prime:$ $\begin{array}{rcl}
        S^\prime\times A^\prime & \to & \Delta_{S^\prime\times\real}\\
        (h_t,a) & \mapsto & (T_t(h_t,a)) \circ (\mathscr{L}_{S,t+1}, 1_\real)
        \end{array}$.
    \end{itemize}
\end{definition}

Note that an MDP is a special case of an NMDP, which naturally induces the non-Markov embedding functor denoted as $\mathbf{N}\in\mton$.

\begin{definition}[\textbf{Non-Markov Embedding}]
 Non-Markov embedding is a functor $\mathbf{N}:\mdpcat\to\nmdpcat$.
For any $\mdp=\mdpdef\in\mdpcat$, there exists $\nmdp^\prime=\mathbf{N}(\mdp)=\nmdpdefp\in\nmdpcat$, $\rho^\prime_0=\rho_0$, $S^\prime=S$, $A^\prime=A$, which satisfies the following property:
    \begin{itemize}
    \item $T_t^\prime:$ $\begin{array}{rcl}
        H_t^\prime\times A^\prime & \to & \Delta_{S^\prime\times\real}\\
        (h_t^\prime,a) & \mapsto & (T_t\circ(\mathscr{L}_{S,t}, 1_A))(h_t^\prime,a)
        \end{array}$.
    \end{itemize}
\end{definition}

Although $\mdpcat$ and $\nmdpcat$ have distinct properties by definition, we discover that they are equivalent categories.
This leads to the following theorem.

\begin{theorem}[\textbf{Equivalence of $\mdpcat$ and $\nmdpcat$}]
~Category $\mdpcat$ and $\nmdpcat$ are equivalent through functor $\mathbf{M}$ and $\mathbf{N}$, i.e.
$\mathbf{M}\circ\mathbf{N}\cong 1_\mdpcat$,~
$\mathbf{N}\circ\mathbf{M}\cong 1_\nmdpcat$.
\end{theorem}

Inspired by the equivalence of $\mdpcat$ and $\nmdpcat$, we believe that by exploring functors in $\mton$, our work will also contribute to the study of algorithms that address non-Markovianity, i.e., identifying more computationally efficient functors in $\funcat(\nmdpcat,\mdpcat)$.

\section{NMDP Constructed from MDP via HAS}
\label{sec:method}
Unlike many algorithmic studies that focus on developing functors in $\ntom$ to handle NMDP's history more effectively, our work focuses on constructing functors in $\mton$ for transforming MDPs into NMDPs.
While $\mathbf{N}$ is such a functor, it is trivial since, in the constructed NMDP, the states prior to the current time are essentially irrelevant to decision-making.
However, our interest lies in functors that can substantially incorporate non-Markovian transition dynamics into MDPs.
To this end, we first introduce the History Aggregator for State (HAS) which enables us to construct such functors.
Similarly, we can also define the History Aggregator for Reward (HAR), which substantially incorporates non-Markovian reward mechanisms into MDPs.
As HAR is not the main content of this paper, we discuss it in the Appendix. Examples and intuitive explanations of the concepts presented in this chapter are also provided in the Appendix.

\begin{definition}[\textbf{History Aggregator for State (HAS)}]
For any MDP $\mdp=\mdpdef$, an HAS $\ha{S}{}\coloneqq\hadef{S}{}$ is a series of maps in which $\mathscr{A}_{S,t}:H_t\to \mathscr{A}(H),~(\forall t\in\nat)$ where $\mathscr{A}(H)$ represents a set called the target state set of $\mathscr{A}_S$.
\end{definition}
As the name implies, the HAS constructs the NMDP state by aggregating the MDP history.
Before detailing our method, we first require that the aggregators be reversible.

\begin{definition}[\textbf{Reversibility of HAS on MDP}]
An HAS $\ha{S}{}=\hadef{S}{}$ on MDP $\mdp=\mdpdef$ is reversible iff there exists a series of maps, denoted as $\ha{S}{\ast}\coloneqq\hadef{S}{\ast}$, which satisfy:
$$\mathscr{A}_{S,t}^\ast(\{\mathscr{A}_{S,\tau}(h_\tau)\}_{\tau=0}^t)=s_t,~~(\forall t\in\nat)$$
where $h_\tau\in H_\tau, ~h_0\prec h_1\prec\cdots\prec h_t, ~s_t=\mathscr{L}_{S,t}(h_t)$.
\end{definition}

The reversibility condition of HAS ensures that a decision-making algorithm can reconstruct the original MDP state from the history of the NMDP.
A reversible HAS-based transformation neither alters the transition dynamics of the original MDP nor increases its inherent complexity.
Consequently, the performance upper bound for the NMDP is identical to that of the original MDP.
Thus, once we know the best achievable performance of the original MDP, we can set the same expectations for the constructed NMDP.
By ensuring that the transformation from the MDP to the NMDP does not increase the inherent difficulty of the original but only requires effective management of the NMDP's history, we can more accurately evaluate a decision-making algorithm's ability to handle non-Markovianity through encoding and memorization of NMDP's history.

Noticing that the latest state extraction operators constitute reversible HAS, we have the following corollary. 

\begin{corollary}
The series of latest state extraction operators, $\mathscr{L}_S\coloneqq\{\mathscr{L}_{S,t}\}_{t=0}^\infty$, constitutes a reversible HAS on MDP with a target state set identical to the state set.
\end{corollary}

When any reversible HAS is applied to an MDP, it generates an NMDP. The NMDP's state set encompasses all aggregated histories of the original MDP.
Its transition dynamics take the NMDP's history as input, decode it into an MDP state, and then apply the MDP's transition dynamics.

\newcommand{\shr}{\!\!\!\!}

\begin{definition}[\textbf{Application of Reversible HAS to MDP}]
\label{def:actionofhas}
The application of a reversible HAS $\ha{S}{}=\hadef{S}{}$ to an MDP $\mdp=\mdpdef$ is an NMDP $\nmdp=\nmdpdefp$, $A^\prime=A$ in which
    $\rho_0^\prime=\rho_0\circ\ha{S,0}{\ast}$;
    $S^\prime=\mathscr{A}(H)$, where $H$ is the history set of $\mdp$, $H^\prime$ is the history set of $\nmdp$;
    $$T_t^\prime:\shr\begin{array}{rcl}H_t^\prime\times A^\prime&\shr\to\shr&\Delta_{S^\prime\times\real} \\
    (h_t^\prime,a)&\shr\mapsto\shr&((T_t\circ(\ha{S,t}{\ast}\circ\mathscr{E}_{S^\prime}, 1_{A}))(h_t^\prime,a))\circ\left(G_{S,h_t^\prime}, 1_\real\right)
    \end{array}$$
    where
    $G_{S,h_t^\prime}: \begin{array}{rcl}S^\prime & \to & S\\
    s^\prime_{t+1} & \mapsto & \ha{S,t+1}{\ast}\left(\mathscr{E}_{S^\prime}(h_t^\prime), s^\prime_{t+1}\right) 
    \end{array}$.
\end{definition}

\textbf{The decoding process, denoted as $\ha{S,t}{\ast}$ is the source of non-Markovianity.} This is because, in the general case, decision-making algorithms have to remember the NMDP's entire history to decode the MDP's current state.

Although reversible HAS naturally induces functors in $\mton$, the definition of reversible HAS alone does not directly produce implementations, necessitating further specification.

For simplicity and practicality in implementation, we restrict our construction to using states alone, not the entire history.
Given this constraint, we examine two types of HAS constructions: one using just states, and the other relying on auxiliary sequences.

\subsection{HAS Constructed Using Just States}
One straightforward way to construct HAS using just states is introducing a binary operator on the state set.

\begin{definition}[\textbf{HAS Induced by Binary Operator}]
The HAS $\hb{S}{}\coloneqq\hbdef{S}{}$ on MDP $\mdp=\mdpdef$ induced by a binary operator $\otimes: S\times S\to S$ is:
$$\hb{S,t}{}: \begin{array}{rcl}
H_t & \to & S \\
h_t & \mapsto & \bigotimes_{\tau=0}^t s_\tau
\end{array}$$
where $\{s_\tau\}_{\tau=0}^t=\mathscr{E}_S(h_t)$, $\bigotimes_{\tau=0}^0 s_\tau\coloneqq s_0$, and $\bigotimes_{\tau=0}^{k+1} s_\tau\coloneqq(\bigotimes_{\tau=0}^k s_\tau)\otimes s_{k+1}, ~~(\forall k\in[0,t)\cap\nat)$.
\end{definition}

According to the above definition, for the HAS induced by operator $\otimes$ to be reversible, it is sufficient for $\otimes$ to be a group operator, meaning $(S,\otimes)$ forms a group.
This approach is quite general since any state set can be extended to form a group, like a free group.
Moreover, in practical problems, states are typically represented as vectors in $\real^n$, and $\real^n$ forms a group under vector addition.

\begin{theorem}[\textbf{Reversibility of HAS Induced by Group Operator}]
\label{th:revgrp}
The HAS $\mathscr{G}_S$ on MDP $\mdp=\mdpdef$ induced by binary operator $\otimes$ is reversible if $(S,\otimes)$ is a group.
\end{theorem}

Let $\mathbf{G}\in\mton$ denote the functor induced by $\mathscr{G}_S$.
The proof of Theorem~\ref{th:revgrp} in the Appendix shows that $\mathbf{G}$ generates an NMDP.
Unlike non-Markov embedding, the HAS $\mathscr{G}_S$ does introduce non-Markovianity.
It transforms an MDP, where decisions rely only on the current state, into an NMDP that requires knowledge of both the current and previous one-step state for decision-making.
However, this state dependency structure is inadequate. 
Later, we'll show that repeated application of this HAS can yield a more complex state dependency structure.

\begin{definition}[\textbf{State Dependency Structure of NMDP History}]
\label{def:statedependency}
The state dependency structure $D_{h_t}$ of a history $h_t=(s_{0:t},a_{0:t-1},r_{0:t-1})$ of NMDP $\nmdp=\nmdpdef$ is a subset of $\nat$
$$D_{h_t}\coloneqq\left\{i\mid\exists s\in S,~\exists a\in A, ~T_t(\sigma_{i}(h_t,s),a)\neq T_t(h_t,a)\right\}$$
where $\sigma_i(h_t,s)=((s_{0:i-1},s,s_{i+1:t}),a_{0:t-1},r_{0:t-1})$.
\end{definition}

According to the above definition, $i\in D_{h_t}$ implies that $s_i$ has an impact on the state transition. Replacing it leads to a change in the subsequent transition probabilities.
The concept of state dependency structure can also be extended to MDPs by defining it in terms of the state dependency structure of the MDP's non-Markov embedding. It has been observed that applying $\mathbf{G}$ once to an MDP can increase the number of aggregated histories required to reconstruct the original MDP state in the resulting NMDP, thereby expanding the cardinality of the state dependency structure. Consequently, we explore applying $\mathbf{G}$ multiple times to the MDP to generate NMDPs with more complex temporal dependencies. To facilitate the application of reversible HAS to NMDPs, and without significant loss of generality, we assume that the transition dynamics of the NMDP depend solely on the state component of the history.
The extension required to apply the reversible HAS to the NMDP is presented in the Appendix. With a slight abuse of notation, we let $\mathbf{G}\in\nton$ also denote the functor induced by HAS.

\begin{theorem}[\textbf{Impact of $\mathbf{G}$ on State Dependency Structure}]
\label{th:gondependency}
    For any MDP $\mdp=\mdpdef$, if it is non-degenerate, then the functor $\mathbf{G}$ induced by some group operator on $S$ ensures that the dependency structure $D_{h^n_t}$ of any history $h^n_t\in H_t^n$ of $\mathbf{G}^n(\mdp)$ satisfies $D_{h^n_t}=[t-n, t]\cap\nat$.
\end{theorem}

The significance of Theorem~\ref{th:gondependency} lies in its provision of a general method for designing a transition mechanism for an NMDP that relies on the states of the previous $n+1$ steps.
Specifically, it is sufficient to apply $\mathbf{G}$ functor $n$ times to a non-degenerate MDP.

\subsection{HAS Constructed Using Auxiliary Sequences}
Another approach to aggregating states is using auxiliary sequences to assign importance weights to these states.
In this way, a binary operator is required to connect an auxiliary sequence with the sequence of states. 

\begin{definition}[\textbf{HAS Induced by Auxiliary Sequences and Binary Operator}]
    The HAS $\mathscr{W}_{S}\coloneqq\{\mathscr{W}_{S,t}\}_{t=0}^\infty$ on MDP $\mdp=\mdpdef$ induced by a series of auxiliary sequences $W\coloneqq\{W_t\}_{t=0}^\infty\coloneqq\{\{w^t_\tau\}_{\tau=0}^t\}_{t=0}^\infty$ and 
    a binary operator $\ast: W\times \mathscr{E}_S(H)\to S$ is:
    $$\mathscr{W}_{S,t}:\begin{array}{rcl}
    H_t & \to & S \\
    h_t & \mapsto & \{w_\tau^t\}_{\tau=0}^t \ast \{s_\tau\}_{\tau=0}^t
    \end{array}$$
    where $\{s_\tau\}_{\tau=0}^t=\mathscr{E}_S(h_t)$.
\end{definition}
The binary operator $\ast$ does not specify how the auxiliary sequence elements interact with the state sequence elements.
Therefore, we define two operators to determine this interaction, one operator $\cdot:\bigsqcup_{t=0}^\infty W_t\times S\to S$ to apply an item in the auxiliary sequence to a state, the other operator $\oplus: S\times S\to S$ to aggregate two states.
Without too much loss of generality, we restrict our discussion to left R-modules, where $(S,\oplus)$ forms an Abelian group, $\cdot$ is a scalar multiplication operator, and $\bigsqcup_{t=0}^\infty W_t$ becomes a unital ring with operators defined. We use $\mathscr{R}_S$ to denote such an HAS. 

Additionally, the correspondence between the elements of the auxiliary sequence and the elements of the state sequence needs to be specified.
For the sake of simplicity in the subsequent discussion, we consider only two types of correspondence, where the auxiliary sequences are prefixes of a given sequence $\{w_\tau\}_{\tau=0}^\infty$: 
$\begin{array}{rccl}\raast:&(\{w_\tau\}_{\tau=0}^t,\{s_\tau\}_{\tau=0}^t)&\mapsto&\bigoplus_{\tau=0}^t(w_\tau\cdot s_{\tau})\\
\laast:&(\{w_\tau\}_{\tau=0}^t,\{s_\tau\}_{\tau=0}^t)&\mapsto&\bigoplus_{\tau=0}^t(w_\tau\cdot s_{t-\tau})
\end{array}
$.
where $\raast$ is referred to as correlation operator, and $\laast$ is referred to as convolution operator.

We examine the reversibility of HAS in these two cases separately.
For the correlation operator case, we have the following theorem:

\begin{theorem}[\textbf{Reversibility of HAS induced by Auxiliary Sequence and Correlation Operator}]
The HAS $\mathscr{R}_S\coloneqq\{\mathscr{R}_{S,t}\}_{t=0}^\infty$ induced by prefixes of auxiliary sequence $\{w_t\}_{t=0}^\infty$ and operator $\raast$ is reversible if each $w_t$ is invertible in the ring.
\end{theorem}

Based on the proof in the Appendix, it is obvious that the correlation operator only results in each set of the dependency structure of the constructed NMDP having a potential of no more than $2$, which is similar to the effect of a group operator.
Therefore, we focus on discussing the properties of the convolution operator.

For the convolution operator case, we have the following theorem:
\begin{theorem}[\textbf{Reversibility of HAS induced by Auxiliary Sequence and Convolution Operator}]
The HAS $\mathscr{R}_S\coloneqq\{\mathscr{R}_{S,t}\}_{t=0}^\infty$ induced by prefixes of auxiliary sequence $\{w_t\}_{t=0}^\infty$ and operator $\laast$ is reversible if $w_0$ is invertible in the ring.
\end{theorem}

We use $\mathbf{R}\in\mton$ to denote the functor induced by reversible $\mathscr{R}_S$ and operator $\laast$, then we have the following theorem about its impact on the state dependency structure of the NMDP.

\begin{theorem}[\textbf{Impact of $\mathbf{R}$ on State Dependency Structure}]
\label{th:rondependency}
For any MDP $\mdp=\mdpdef$, if it is non-degenerate, then the functor $\mathbf{R}$ induced by prefixes of auxiliary sequence $\{w_\tau\}_{\tau=0}^\infty$ and operator $\laast$ ensures that the dependency structure $D_{h^\prime_t}$ of any history $h^\prime_t\in H^\prime_t$ of $\mathbf{R}(\mdp)$ satisfies $D_{h^\prime_t}=\{t-\tau\mid(\mathbf{w}^{-1})_{0,\tau}\neq0\}$, where $\mathbf{w}^{-1}$ is the inverse matrix of $\mathbf{w}$ in the ring, $0$ is the zero element of the ring.
\end{theorem}

The proof of Theorem~\ref{th:rondependency} indicates that it is possible to achieve any specific state dependency structure in the history of the constructed NMDP by modifying the elements of the upper triangular band matrix $\mathbf{w}^{-1}$.
This provides a general method for designing NMDPs with desired state dependency structures.

\section{Experiments}
\label{sec:exp}

Although the main contribution of this paper is theoretical, we have designed and implemented several highly versatile environment wrappers based on the two methods proposed in Chapter \ref{sec:method} for constructing functors in $\mton$: the group-operator-based and the convolution-based method.
These wrappers are implemented using a computationally efficient approach with a time complexity of $O(n)$ and can be applied to environments that conform to the Gymnasium \citep{gym} interface.

We use the PPO algorithm implemented in the Stable-Baselines3 library \citep{sb3} and the LSTM-PPO algorithm of the Stable-Baselines3-Contrib library \citep{sb3contrib} as standard reinforcement learning algorithms.
These algorithms are tested in Gymnasium environments wrapped with our wrappers, with reinforcement learning training experiments conducted using the framework provided by the RL-Baselines3-Zoo library \citep{rlzoo}.

We assume $s_{-1}\coloneqq\mathbf{0}$, $\lambda\in[0,1]$, $\exists k\forall t, ~s_t\in\real^k$, and employ the following functor to implement wrappers:
\begin{itemize}
\item $\mathbf{S}$ induced by HAS $\mathscr{S}_{S,t}: h_t\mapsto\sum_{\tau=0}^t s_\tau$;
\item $\mathbf{D}$ induced by HAS $\mathscr{D}_{S,t}: h_t\mapsto s_t-s_{t-1}$;
\item $\mathbf{S}_\lambda$ induced by HAS $\mathscr{S}^\lambda_{S,t}: h_t\mapsto\sum_{\tau=0}^t\lambda^\tau s_{t-\tau}$;
\item $\mathbf{D}_\lambda$ induced by HAS $\mathscr{D}^\lambda_{S,t}: h_t\mapsto s_t-\lambda s_{t-1}$.
\end{itemize}

From the definitions of $\mathbf{S}$, $\mathbf{D}$, $\mathbf{S}_\lambda$, $\mathbf{D}_\lambda$, we can derive the state dependency structure of histories in the NMDP under their application.
$\mathbf{S}$, $\mathbf{S}_\lambda$ introduce dependencies of $s_t$, $s_{t-1}$, and $\mathbf{D}$, $\mathbf{D}_\lambda$ introduce dependencies of $s_{0:t}$. 
The state dependency weights induced by $\mathbf{S}_\lambda$ and $\mathbf{D}_\lambda$ exhibit an exponential decay with respect to the temporal distance from the current time step, governed by the factor $\lambda$, while $\mathbf{S}$ and $\mathbf{D}$ introduce uniformly distributed dependency.

Our experiments use CartPole-v1 and Pendulum-v1 as the base MDP environments, and we employ the following functors to obtain the NMDP environments:
\begin{center}$\mathbf{S}^n$, $\mathbf{D}^n$ for $n\in\{0,1,2,3,4,5\}$\end{center}
\begin{center}$\mathbf{S}_\lambda$, $\mathbf{D}_\lambda$ for $\lambda\in\{\frac{0}{5}, \frac{1}{5},\frac{2}{5},\frac{3}{5},\frac{4}{5},\frac{5}{5}\}$\end{center}

All hyperparameters required for training were kept at their default settings from the RL-Baselines3-Zoo library, and the training process was initiated using the library's provided command-line instructions. Each combination of environment, wrapper, and algorithm was trained $3$ times, $10^6$ timesteps for each time, evaluated every $2\times10^5$ timesteps, and the checkpoints with the best average episode reward were selected.

\begin{figure}[h!]
  \centering
  \subfigure{\includegraphics[width=0.23\textwidth]{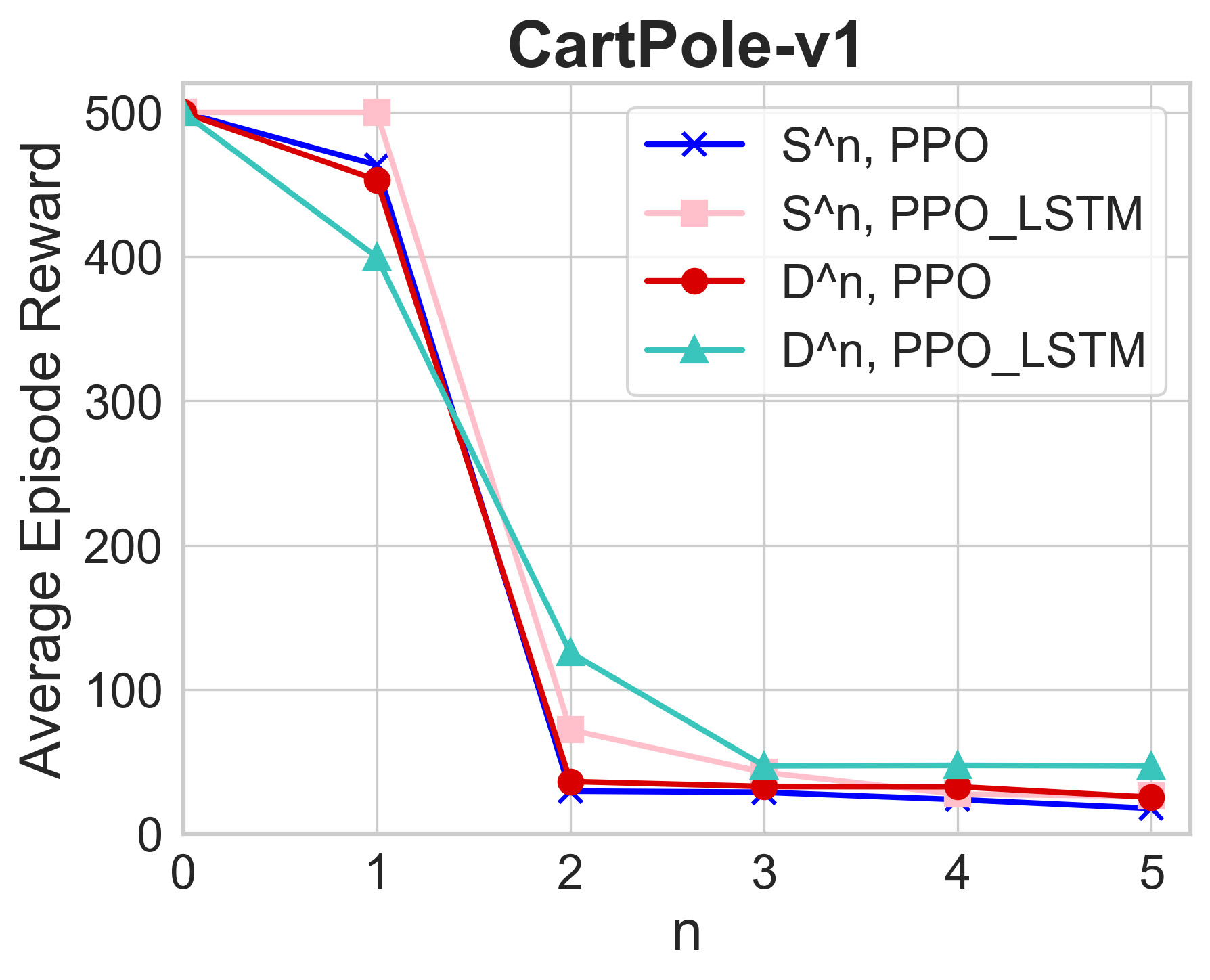}}
  \subfigure{\includegraphics[width=0.23\textwidth]{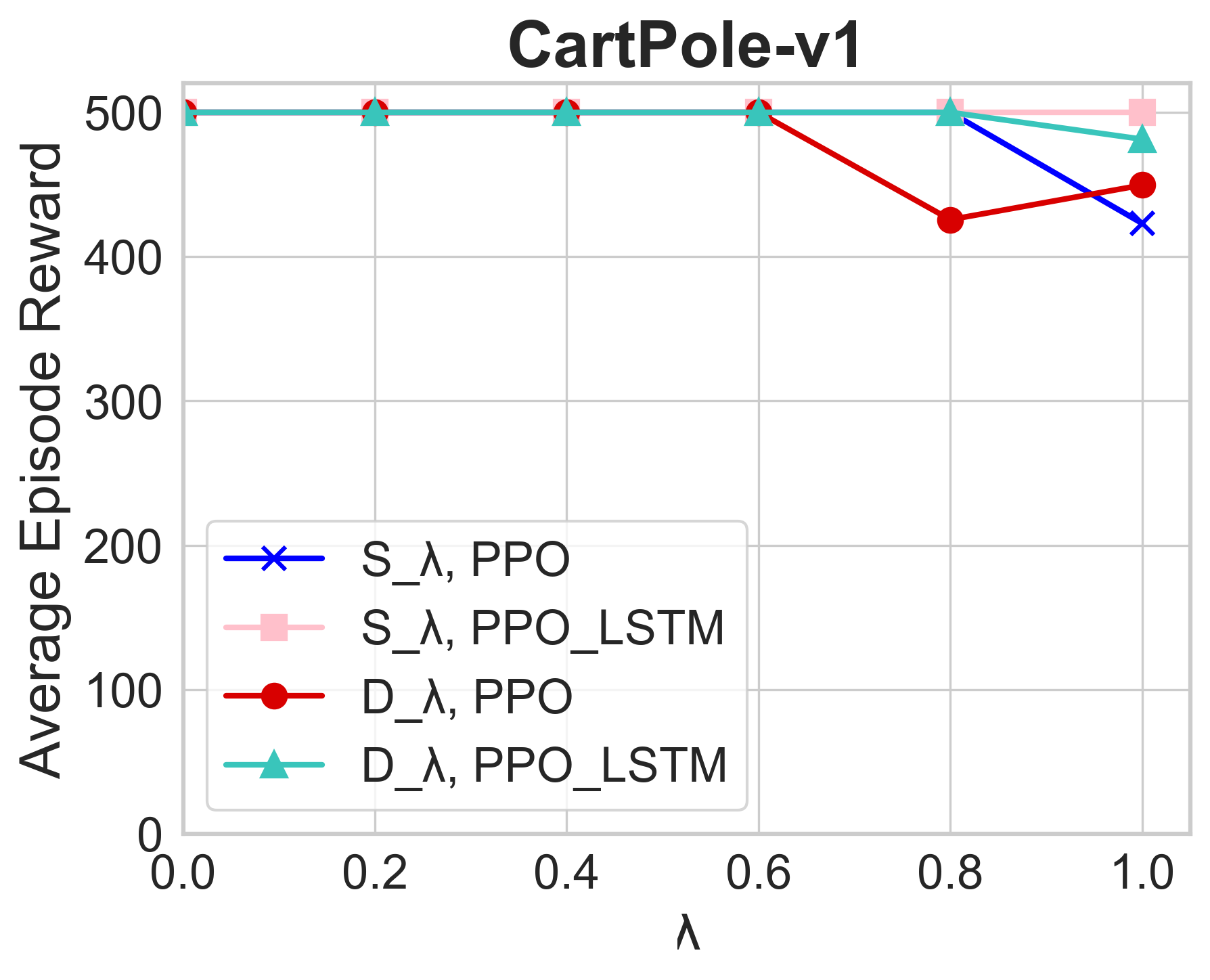}}
  \vskip\baselineskip
  \subfigure{\includegraphics[width=0.23\textwidth]{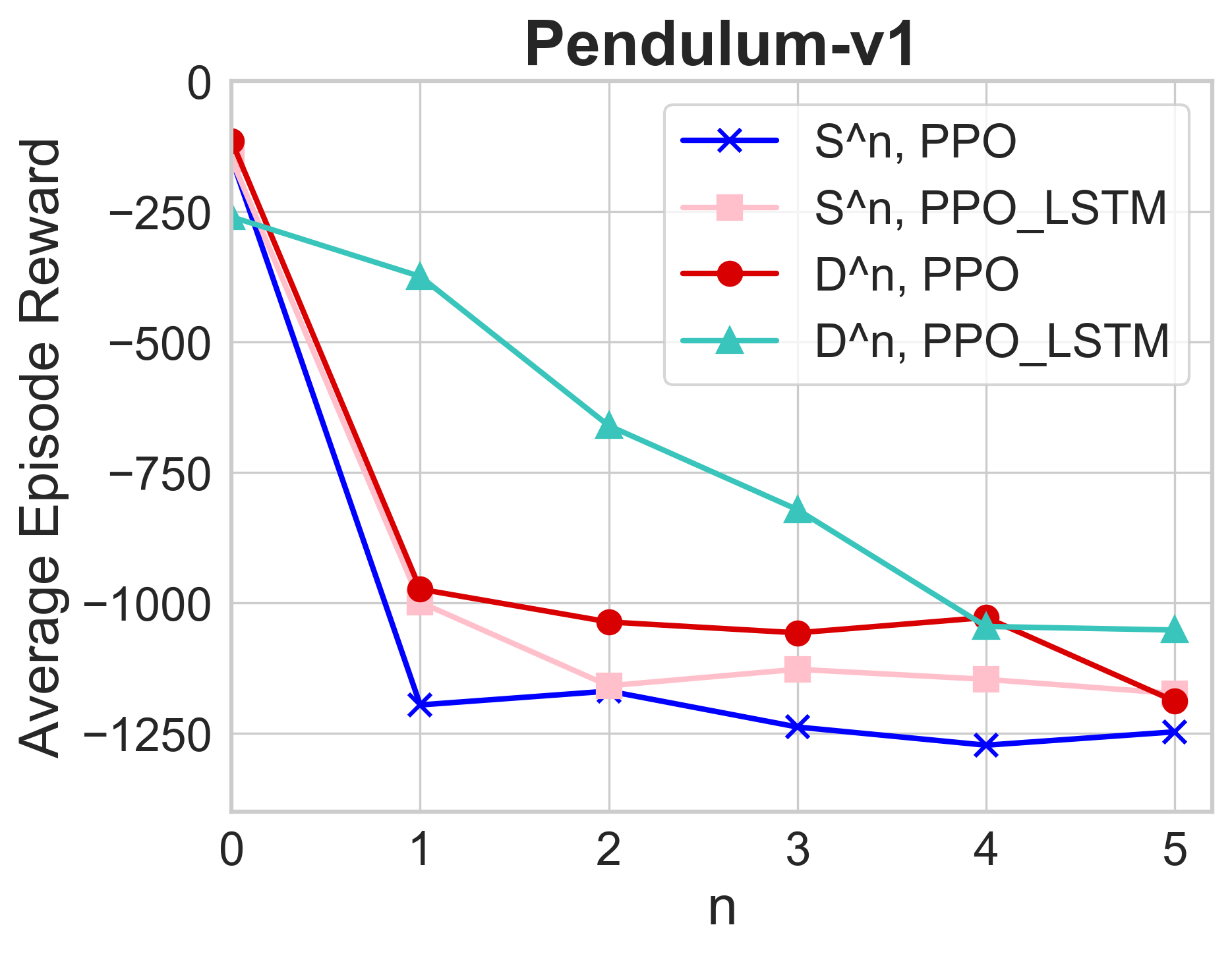}}
  \subfigure{\includegraphics[width=0.23\textwidth]{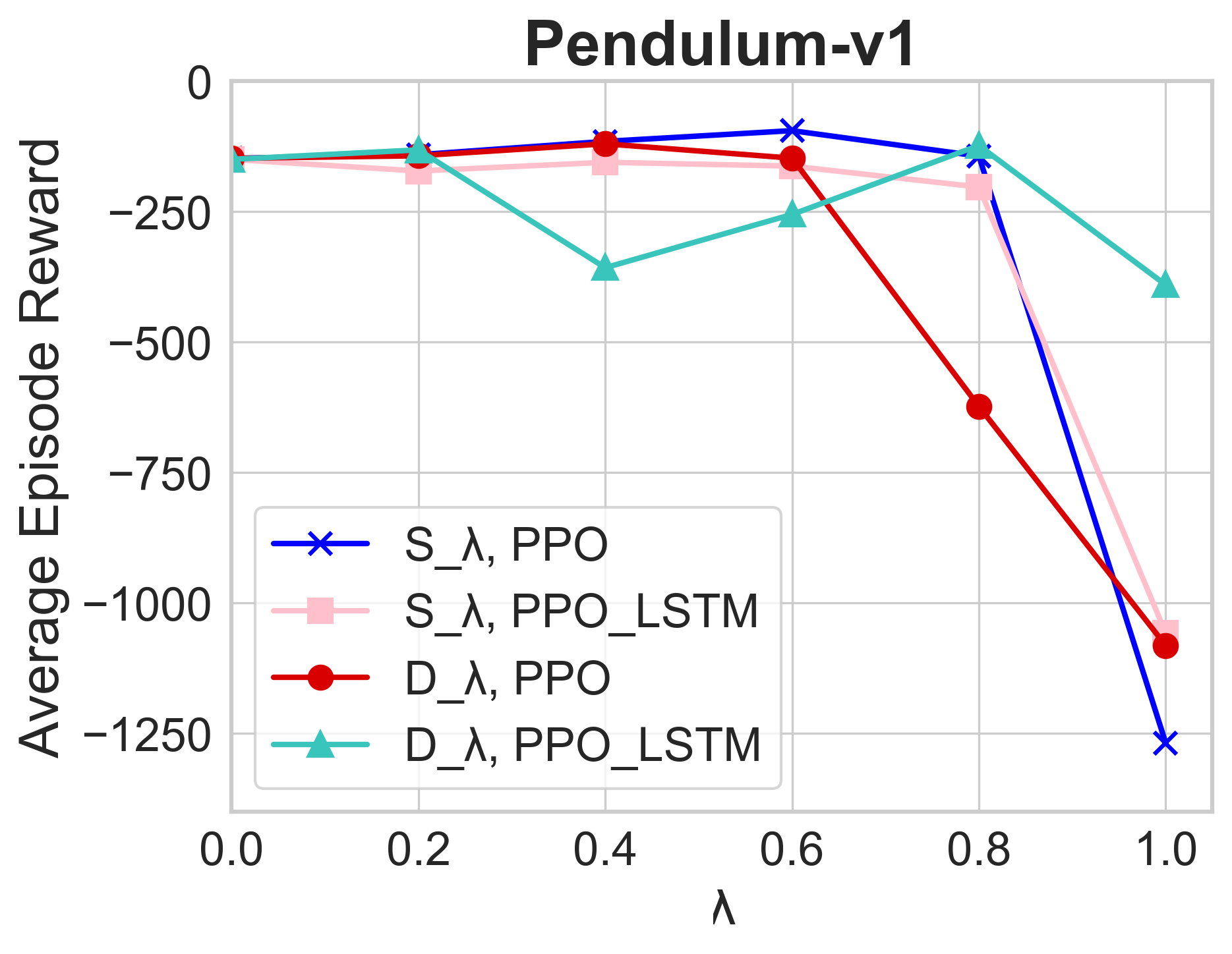}}
  \caption{Experimental results for each combination of environment, algorithm, and wrapper. The x-axis represents the wrapper parameter, while the y-axis shows the average episode reward. Lines of different colors indicate different combinations of algorithms and wrapper types.}
  \label{fig:exp_result}
\end{figure}

Theoretically, an increase in $n$ of $\mathbf{S}^n$ and $\mathbf{D}^n$ or $\lambda$ of $\mathbf{S}_\lambda$ and $\mathbf{D}_\lambda$ makes the non-Markov environment more challenging for RL algorithms to solve, as it introduces more complex temporal dependencies.
As shown in \ref{fig:exp_result}, the results of our experiments also indicate this property: the average episode reward, plotted as a function of $n$ or $\lambda$, generally shows a declining trend for each combination of environment, algorithm, and wrapper.

Another conclusion drawn from the experimental results is that, for any given combination of environment and algorithm, the performance degradation caused by increasing $\lambda$ from $0$ to $1$ is comparable to the degradation resulting from increasing $n$ from $0$ to $1$. This observation is consistent with our theoretical analysis, which shows that increasing $\lambda$ from $0$ to $1$ progressively introduces a greater dependence on previous states, eventually matching the dependence introduced by a wrapper with $n=1$.

By comparing the performance of the PPO and LSTM-PPO algorithms in environments wrapped with varying degrees of dependency complexity, we observe that LSTM-PPO outperforms PPO when the dependencies are more complex.
This is due to LSTM-PPO's ability to recall past states, which also aligns with our intuition.

\section{Summary and Future Work}
\label{sec:summary}
We propose a general and effective method to construct NMDPs from MDPs using techniques from category theory and algebra.
Our method provides a comprehensive way to evaluate decision-making, especially RL algorithms' ability to handle non-Markovianity through encoding and memorization.

Theoretical analysis demonstrates that our method possesses strong expressive power and can represent a wide variety of temporal state dependency structures. Using this theoretical framework, we transformed classical Markovian environments into non-Markovian ones and tested them with both PPO algorithms, with and without LSTM. This validation confirms our method's effectiveness in introducing varying degrees of non-Markovian characteristics into Markovian environments.

The examples tested in our experiments represent only a small subset of the cases within the theoretical framework. The functor forms that exhibit desirable properties through our theory are highly diverse, extending beyond those tested. This is due to our proofs being conducted within abstract algebraic structures, which encompass numerous notable specific cases.

Our study also has limitations that warrant further investigation. For instance, we did not explore the introduction of randomness into HAS using random sequences or mappings. Additionally, we did not address the theory of HAR, which remains relevant for reward shaping. Moreover, our theoretical analysis is restricted to reversible HAS, and we have not developed a theory for non-reversible cases. For example, a common method to create non-Markovian environments is by obscuring information in the observations of an MDP to transform it into a POMDP. Combining this method with our approach and exploring their combined expressive power presents a promising avenue for future research.

\section*{Ethical Statement}

There are no ethical issues.

\section*{Acknowledgments}
The authors have no acknowledgments to report.
\bibliographystyle{named}
\bibliography{ijcai25}

\end{document}


\maketitle

\maketitle

\textbf{Code}: Considering anonymity, the code on Github will be made public after this article is accepted.

\newcommand{\funcat}{\mathbf{Fun}}
\newcommand{\mton}{\funcat(\mdpcat,\nmdpcat)}
\newcommand{\ntom}{\funcat(\nmdpcat,\mdpcat)}
\newcommand{\nton}{\funcat(\nmdpcat,\nmdpcat)}
\newcommand{\nat}{\mathbb{N}}
\newcommand{\natplus}{\nat^{+}}
\newcommand{\real}{\mathbb{R}}
\newcommand{\mdp}{\mathcal{M}}
\newcommand{\nmdp}{\mathcal{N}}
\newcommand{\mdpcat}{\mathfrak{M}}
\newcommand{\nmdpcat}{\mathfrak{N}}
\newcommand{\mdpdef}{\langle\rho_0, S, A, \{T_t\}_{t=0}^\infty\rangle}
\newcommand{\mdpdefp}{\langle\rho_0^\prime, S^\prime, A^\prime, \{T^\prime_t\}_{t=0}^\infty\rangle}
\newcommand{\nmdpdef}{\langle\rho_0, S, A, \{T_t\}_{t=0}^\infty\rangle}
\newcommand{\nmdpdefp}{\langle\rho_0^\prime, S^\prime, A^\prime, \{T^\prime_t\}_{t=0}^\infty\rangle}
\newcommand{\morphism}[1]{(#1_{S},#1_{A},#1_{\real})}
\newcommand{\ha}[2]{\mathscr{A}_{#1}^{#2}}
\newcommand{\hadef}[2]{\{\mathscr{A}_{#1,t}^{#2}\}_{t=0}^\infty}
\newcommand{\hb}[2]{\mathscr{B}_{#1}^{#2}}
\newcommand{\hbdef}[2]{\{\mathscr{B}_{#1,t}^{#2}\}_{t=0}^\infty}
\newcommand{\laast}{\overset{\scriptscriptstyle\leftarrow}{\ast}}
\newcommand{\raast}{\overset{\scriptscriptstyle\rightarrow}{\ast}}
\newcommand{\shr}{\!\!\!\!}
\newtheorem{lemma}{Lemma}
\newcommand{\nmdpdefj}{\langle\rho_0,S,A,\{J_t\}_{t=0}^\infty\rangle}
\newcommand{\nmdpdefjp}{\langle\rho_0^\prime,S^\prime,A^\prime,\{J_t^\prime\}_{t=0}^\infty\rangle}

\newcommand{\mdpdefh}{\langle\rho_0,H,A,\{T_t\}_{t=0}^\infty\rangle}
\newcommand{\mdpdefhp}{\langle\rho_0^\prime,H^\prime,A^\prime,\{T_t^\prime\}_{t=0}^\infty\rangle}
\newcommand{\dar}[2]{\overset{#1}{\underset{#2}{\rightrightarrows}}}

\section{Proof of Well-Definedness of MDP Category}
\begin{lemma}
MDP Category $\mdpcat$ is well-defined.
\end{lemma}
\begin{proof}
It is sufficient to prove that the composition of morphisms remains a morphism.\\
For any MDP $\mdp,\mdp^\prime,\mdp^{\prime\prime}\in\mdpcat$ and any morphism: $\phi:\mdp\to\mdp^\prime$, $\phi^\prime:\mdp^\prime\to\mdp^{\prime\prime}$, by definition of composition of MDP category we have
$$\phi^\prime\circ\phi=(\phi_S^\prime\circ\phi_S,\phi_A^\prime\circ\phi_A,\phi_\real^\prime\circ\phi_\real)$$
Since
$$\rho_0=\rho_0^\prime\circ\phi_S$$
$$\rho^\prime_0=\rho^{\prime\prime}_0\circ\phi^\prime_S$$
we have
$$\rho_0=(\rho^{\prime\prime}_0\circ\phi^\prime_S)\circ\phi_S=\rho^{\prime\prime}_0\circ(\phi^\prime_S\circ\phi_S)=\rho^{\prime\prime}_0\circ(\phi^\prime\circ\phi)_S$$
similarly, since
$$T_t(s,a)=((T^\prime_t\circ(\phi_S,\phi_A))(s,a))\circ(\phi_S,\phi_\real)$$
$$T^\prime_t(s^\prime,a^\prime)=((T^{\prime\prime}_t\circ(\phi^\prime_S,\phi^\prime_A))(s^\prime,a^\prime))\circ(\phi^\prime_S,\phi^\prime_\real)$$
take $(s^\prime,a^\prime)=(\phi_S,\phi_A)(s,a)$, we have
$$T_t(s,a)=((T^\prime_t\circ(\phi_S,\phi_A))(s,a))\circ(\phi_S,\phi_\real)$$
$$=(T^\prime_t((\phi_S,\phi_A)(s,a)))\circ(\phi_S,\phi_\real)$$
$$=(T^\prime_t(s^\prime,a^\prime))\circ(\phi_S,\phi_\real)$$
$$=(((T^{\prime\prime}_t\circ(\phi^\prime_S,\phi^\prime_A))(s^\prime,a^\prime))\circ(\phi^\prime_S,\phi^\prime_\real))\circ(\phi_S,\phi_\real)$$
$$=((T^{\prime\prime}_t\circ(\phi^\prime_S,\phi^\prime_A))(s^\prime,a^\prime))\circ((\phi^\prime\circ\phi)_S,(\phi^\prime\circ\phi)_\real)$$
$$=((T^{\prime\prime}_t\circ(\phi^\prime_S,\phi^\prime_A))((\phi_S,\phi_A)(s,a)))\circ((\phi^\prime\circ\phi)_S,(\phi^\prime\circ\phi)_\real)$$
$$=((T^{\prime\prime}_t\circ((\phi^\prime\circ\phi)_S,(\phi^\prime\circ\phi)_A))(s,a))\circ((\phi^\prime\circ\phi)_S,(\phi^\prime\circ\phi)_\real)$$
therefore $\phi^\prime\circ\phi$ is still a morphism, i.e. $\mdpcat$ is a well-defined category.
\end{proof}

\section{Proof of Well-Definedness of NMDP Category}
\begin{lemma}
NMDP Category $\nmdpcat$ is well-defined.
\end{lemma}
\begin{proof}
It is sufficient to prove that the composition of morphisms remains a morphism.\\
For any NMDP $\nmdp,\nmdp^\prime,\nmdp^{\prime\prime}\in\nmdpcat$ and any morphism: $\psi:\nmdp\to\nmdp^\prime$, $\psi^\prime:\nmdp^\prime\to\nmdp^{\prime\prime}$, by definition of composition of NMDP category we have
$$\psi^\prime\circ\psi=(\psi_S^\prime\circ\psi_S,\psi_A^\prime\circ\psi_A,\psi_\real^\prime\circ\psi_\real)$$
Since
$$\rho_0=\rho_0^\prime\circ\psi_S$$
$$\rho^\prime_0=\rho^{\prime\prime}_0\circ\psi^\prime_S$$
we have
$$\rho_0=(\rho^{\prime\prime}_0\circ\psi^\prime_S)\circ\psi_S=\rho^{\prime\prime}_0\circ(\psi^\prime_S\circ\psi_S)=\rho^{\prime\prime}_0\circ(\psi^\prime\circ\psi)_S$$
similarly, we have
$$T_t(h_t,a)=((T_t^\prime\circ(\psi_{H_t},\psi_A))(h_t,a))\circ(\psi_S,\psi_\real)$$
$$T^\prime_t(h_t^\prime,a^\prime)=((T_t^{\prime\prime}\circ(\psi^\prime_{H_t},\psi^\prime_A))(h^\prime_t,a^\prime))\circ(\psi^\prime_S,\psi^\prime_\real)$$
take $(h^\prime_t,a^\prime)=(\psi_{H_t},\psi_A)(h_t,a)$, we have
$$T_t(h_t,a)=((T^\prime_t\circ(\psi_{H_t},\psi_A))(h_t,a))\circ(\psi_S,\psi_\real)$$
$$=(T^\prime_t((\psi_{H_t},\psi_A)(h_t,a)))\circ(\psi_S,\psi_\real)$$
$$=(T^\prime_t(h_t^\prime,a^\prime))\circ(\psi_S,\psi_\real)$$
$$=(((T^{\prime\prime}_t\circ(\psi^\prime_{H_t},\psi^\prime_A))(h_t^\prime,a^\prime))\circ(\psi^\prime_S,\psi^\prime_\real))\circ(\psi_S,\psi_\real)$$
$$=((T^{\prime\prime}_t\circ(\psi^\prime_{H_t},\psi^\prime_A))(h_t^\prime,a^\prime))\circ((\psi^\prime\circ\psi)_S,(\psi^\prime\circ\psi)_\real)$$
$$=((T^{\prime\prime}_t\circ(\psi^\prime_{H_t},\psi^\prime_A))((\psi_{H_t},\psi_A)(h_t,a)))\circ((\psi^\prime\circ\psi)_S,(\psi^\prime\circ\psi)_\real)$$
$$=((T^{\prime\prime}_t\circ(\psi_{H_t}^\prime\circ\psi_{H_t},(\psi^\prime\circ\psi)_A))(h_t,a))\circ((\psi^\prime\circ\psi)_S,(\psi^\prime\circ\psi)_\real)$$
since
$$\psi_{H_t}=(\psi_S^{\langle t+1\rangle},\psi_A^{
        \langle t\rangle},\psi_\real^{\langle t\rangle})$$
$$\psi^\prime_{H_t}=(\psi_S^{\prime\langle t+1\rangle},\psi_A^{\prime
        \langle t\rangle},\psi_\real^{\prime\langle t\rangle})$$
we have
$$\psi_{H_t}^\prime\circ\psi_{H_t}=(\psi_S^{\prime\langle t+1\rangle},\psi_A^{\prime
        \langle t\rangle},\psi_\real^{\prime\langle t\rangle})\circ(\psi_S^{\langle t+1\rangle},\psi_A^{
        \langle t\rangle},\psi_\real^{\langle t\rangle})$$
$$=(\psi_S^{\prime\langle t+1\rangle}\circ\psi_S^{\langle t+1\rangle},\psi_A^{\prime
        \langle t\rangle}\circ\psi_A^{
        \langle t\rangle},\psi_\real^{\prime\langle t\rangle}\circ\psi_\real^{\langle t\rangle})$$
$$=((\psi^\prime\circ\psi)_S^{\langle t+1\rangle},(\psi^\prime\circ\psi)_A^{
        \langle t\rangle},(\psi^\prime\circ\psi)_\real^{\langle t\rangle})$$
$$=(\psi^\prime\circ\psi)_{H_t}$$
therefore
$$T_t(h_t,a)=((T^{\prime\prime}_t\circ((\psi^\prime\circ\psi)_{H_t},(\psi^\prime\circ\psi)_A))(h_t,a))\circ((\psi^\prime\circ\psi)_S,(\psi^\prime\circ\psi)_\real)$$
which means $\psi^\prime\circ\psi$ is still a morphism, i.e. $\nmdpcat$ is a well-defined category.
\end{proof}
\section{Proof of Well-Definedness of Non-Markov Embedding as a Functor}
\begin{lemma}
Non-Markov embedding $\mathbf{N}$ is a well-defined functor from $\mdpcat$ to $\nmdpcat$ if $\forall\mdp,\mdp^\prime\in\mdpcat,\forall\phi\in\mdpcat(\mdp,\mdp^\prime)$, $\mathbf{N}(\phi)=\phi$.
\end{lemma}
\begin{proof}
For any $\mdp=\mdpdef, \ \mdp^\prime=\mdpdefp, \ \mdp^{\prime\prime}=\langle\rho_0^{\prime\prime},S^{\prime\prime},A^{\prime\prime},\{T_t^{\prime\prime}\}_{t=0}^\infty\rangle\in\mdpcat$ and any morphism $\phi\in\mdpcat(\mdp,\mdp^\prime), \ \phi^\prime\in\mdpcat(\mdp^\prime,\mdp^{\prime\prime})$, 
denote $\nmdp\coloneqq\mathbf{N}(\mdp)=\nmdpdefj$ and $\nmdp^\prime\coloneqq\mathbf{N}(\mdp^\prime)=\nmdpdefjp$.\\\\
If the condition $\forall\phi, \ \mathbf{N}(\phi)=\phi$ holds, then
$$\mathbf{N}(1_\mdp)=1_\mdp=(1_S,1_A,1_\real)=1_{\mathbf{N}(\mdp)}$$
and 
$$\mathbf{N}(\phi^\prime\circ\phi)=\phi^\prime\circ\phi=\mathbf{N}(\phi^\prime)\circ\mathbf{N}(\phi)$$
which means the identity morphisms and associativity of morphisms in $\mdpcat$ is preserved by $\mathbf{N}$. \\
Therefore, for the well-definedness of $\mathbf{N}$, we only need to prove that the condition $\forall\phi, \ \mathbf{N}(\phi)=\phi$ is compatible with the action of $\mathbf{N}$ on $ob(\mdpcat)$ which is given by the definition of $\mathbf{N}$, i.e. $\mathbf{N}(\phi)=\phi\in\nmdpcat(\nmdp,\nmdp^\prime)$.\\
Since $\mathbf{N}$ does nothing to $\rho_0,\rho_0^\prime;S,S^\prime;A,A^\prime$, we only need to prove that the condition is compatible with the action of $\mathbf{N}$ on $T_t,T^\prime_t$.\\
By the definition of morphisms in $\mdpcat$, we have
$$T_t(s,a)=((T_t^\prime\circ(\phi_S,\phi_A))(s,a))\circ(\phi_S,\phi_\real)$$
By the definition of $\mathbf{N}$, we have
$$J_t(h_t,a)=(T_t\circ(\mathscr{L}_{S,t},1_A))(h_t,a)$$
$$J_t^\prime(h^\prime_t,a^\prime)=(T_t^\prime\circ(\mathscr{L}_{S^\prime,t},1_{A^\prime}))(h^\prime_t,a^\prime)$$
denote $s\coloneqq\mathscr{L}_{S,t}(h_t),s^\prime\coloneqq\mathscr{L}_{S^\prime,t}(h_t^\prime)$ then
$$J_t(h_t,a)=(T_t\circ(\mathscr{L}_{S,t},1_A))(h_t,a)$$
$$=T_t((\mathscr{L}_{S,t},1_A)(h_t,a))$$
$$=T_t(s,a)$$
$$=((T_t^\prime\circ(\phi_S,\phi_A))(s,a))\circ(\phi_S,\phi_\real)$$
$$=(T_t^\prime((\phi_S,\phi_A)(s,a)))\circ(\phi_S,\phi_\real)$$
$$=(T_t^\prime(s^\prime,a^\prime))\circ(\phi_S,\phi_\real)$$
$$=(T_t^\prime((\mathscr{L}_{S^\prime,t},1_{A^\prime})(h_t^\prime,a^\prime)))\circ(\phi_S,\phi_\real)$$
$$=((T_t^\prime\circ(\mathscr{L}_{S^\prime,t},1_{A^\prime}))(h_t^\prime,a^\prime))\circ(\phi_S,\phi_\real)$$
$$=(J_t^\prime(h_t^\prime,a^\prime))\circ(\phi_S,\phi_\real)$$
$$=(J_t^\prime((\phi_{H_t},\phi_A)(h_t,a)))\circ(\phi_S,\phi_\real)$$
$$=((J_t^\prime\circ(\phi_{H_t},\phi_A))(h_t,a))\circ(\phi_S,\phi_\real)$$
therefore, by definition of morphisms in $\nmdpcat$, $\phi\in\nmdpcat(\nmdp,\nmdp^\prime)$, which means $\forall\phi, \ \mathbf{N}(\phi)=\phi$ is compatible with the action of $\mathbf{N}$ on $T_t,T_t^\prime$, i.e. $\mathbf{N}$ is a well-defined functor from $\mdpcat$ to $\nmdpcat$.
\end{proof}

\section{Proof of Well-Definedness of Markov Abstraction as a Functor}
\begin{lemma}
Markov abstraction $\mathbf{M}$ is a well-defined functor from $\nmdpcat$ to $\mdpcat$ if $\forall\nmdp,\nmdp^\prime\in\nmdpcat,\forall\psi=(\psi_S,\psi_A,\psi_\real)\in\nmdpcat(\nmdp,\nmdp^\prime)$, $\mathbf{M}(\psi)=(\psi_H,\psi_A,\psi_\real)$, where 
$$H\coloneqq\bigsqcup_{t=0}^\infty H_t=\bigsqcup_{t=0}^\infty S^{t+1}\times A^t\times\real^t$$
$$H^\prime\coloneqq\bigsqcup_{t=0}^\infty H^\prime_t=\bigsqcup_{t=0}^\infty {S^\prime}^{t+1}\times {A^\prime}^t\times\real^t$$

and $\forall t\in\nat$ $$\psi_H:\begin{array}{rcl} H & \to & H^\prime\\ (s_{0:t},a_{0:t-1},r_{0:t-1}) & \mapsto & \psi_{H_t}(s_{0:t},a_{0:t-1},r_{0:t-1})\end{array}$$
\end{lemma}
\begin{proof}
For any $\nmdp=\nmdpdefj, \ \nmdp^\prime=\nmdpdefjp, \ \nmdp^{\prime\prime}=\langle\rho_0^{\prime\prime},S^{\prime\prime},A^{\prime\prime},\{J_t^{\prime\prime}\}_{t=0}^\infty\rangle\in\nmdpcat$ and any morphism $\psi\in\nmdpcat(\nmdp,\nmdp^\prime), \ \psi^\prime\in\nmdpcat(\nmdp^\prime,\nmdp^{\prime\prime})$, 
denote $\mdp\coloneqq\mathbf{M}(\nmdp)=\mdpdefh$ and $\mdp^\prime\coloneqq\mathbf{M}(\nmdp^\prime)=\mdpdefhp$, where $H\coloneqq\bigsqcup_{t=0}^\infty H_t=\bigsqcup_{t=0}^\infty S^{t+1}\times A^t\times\real^t$, $H^\prime\coloneqq\bigsqcup_{t=0}^\infty H^\prime_t=\bigsqcup_{t=0}^\infty {S^{\prime}}^{t+1}\times {A^\prime}^t\times{\real}^t$.\\\\
If the condition $\forall\psi, \ \mathbf{M}(\psi)=(\psi_H,\psi_A,\psi_\real)$ holds, then
$$\mathbf{M}(1_\nmdp)=1_\nmdp=(1_H,1_A,1_\real)=1_{\mathbf{M}(\nmdp)}$$
and 
$$\mathbf{M}(\psi^\prime\circ\psi)=\mathbf{M}((\psi^\prime\circ\psi)_S,(\psi^\prime\circ\psi)_A,(\psi^\prime\circ\psi)_\real)$$
$$=((\psi^\prime\circ\psi)_H,(\psi^\prime\circ\psi)_A,(\psi^\prime\circ\psi)_\real)$$
$$=(\psi_H^\prime\circ\psi_H,\psi_A^\prime\circ\psi_A,\psi_\real^\prime\circ\psi_\real)$$
$$=(\psi^\prime_H,\psi^\prime_A,\psi^\prime_\real)\circ(\psi_H,\psi_A,\psi_\real)$$
$$=\mathbf{M}(\psi^\prime)\circ\mathbf{M}(\psi)$$
which means the identity morphisms and associativity of morphisms in $\nmdpcat$ is preserved by $\mathbf{M}$. \\
Therefore, for the well-definedness of $\mathbf{M}$, we only need to prove that the condition $\forall\psi, \ \mathbf{M}(\psi)=(\psi_H,\psi_A,\psi_\real)$ is compatible with the action of $\mathbf{M}$ on $ob(\nmdpcat)$ which is given by the definition of $\mathbf{M}$, i.e. $\mathbf{M}(\psi)=(\psi_H,\psi_A,\psi_\real)\in\mdpcat(\mdp,\mdp^\prime)$.\\
Since $\mathbf{M}$ does nothing to $\rho_0,\rho_0^\prime;A,A^\prime$, we only need to prove that the condition is compatible with the action of $\mathbf{M}$ on $S,S^\prime;J_t,J^\prime_t$.\\
For any $s\in S$, let $s^\prime=\psi_S(s)\in S^\prime$.
and $h_t=(s_{0:t},a_{0:t-1},r_{0:t-1})\in H$, if $\forall i, s_i^\prime=\psi_S(s_i),a_i^\prime=\psi_A(a_i),r_i^\prime=\psi_\real(r_i)$, then 
$h_t^\prime=(s^\prime_{0:t},a^\prime_{0:t-1},r^\prime_{0:t-1})\in H^\prime$.  \\
We have
$$h_t^\prime=(s^\prime_{0:t},a^\prime_{0:t-1},r^\prime_{0:t-1})$$
$$=(\psi_S^{\langle t+1\rangle}(s_{0:t}),\psi_A^{\langle t\rangle}(a_{0:t-1}),\psi^{\langle t\rangle}_\real(r_{0:t-1}))$$
$$=(\psi_S^{\langle t+1\rangle},\psi_A^{\langle t\rangle},\psi^{\langle t\rangle}_\real)(s_{0:t},a_{0:t-1},r_{0:t-1})$$
$$=\psi_H(h_t)$$
therefore the condition is compatible with the action of $\mathbf{M}$ on $S,S^\prime$. \\
By the definition of morphisms in $\nmdpcat$, we have
$$J_t(h_t,a)=((J^\prime_t\circ(\psi_{H_t},\psi_A))(h_t,a))\circ(\psi_S,\psi_\real)$$
By the definition of $\mathbf{M}$ we have
$$T_t(h_t,a)=(J_t(h_t,a))\circ(\mathscr{L}_{S,t+1},1_\real)$$
$$T^\prime_t(h^\prime_t,a^\prime)=(J^\prime_t(h^\prime_t,a^\prime))\circ(\mathscr{L}_{S^\prime,t+1},1_\real)$$
For any $h_{t+1}\coloneqq(s_{0:t+1},a_{0:t},r_{0:t})\in H_{t+1}\subset H$
, we have
$$(\psi_S\circ\mathscr{L}_{S,t+1})(h_{t+1})=\psi_S(s_{t+1})$$
and
$$(\mathscr{L}_{S^\prime,t+1}\circ\psi_H)(h_{t+1})=\mathscr{L}_{S^\prime,t+1}(\psi_{H_{t+1}}(h_{t+1}))$$
$$=\mathscr{L}_{S^\prime,t+1}(\psi_S^{\langle t+2\rangle}(s_{0:t+1}),\psi_A^{\langle t+1\rangle}(a_{0:t}),\psi_\real^{\langle t+1\rangle}(r_{0:t}))$$
$$=\psi_S(s_{t+1})$$
thus $\psi_S\circ\mathscr{L}_{S,t+1}=\mathscr{L}_{S^\prime,t+1}\circ\psi_H$, which implies that
$$T_t(h_t,a)=(J_t(h_t,a))\circ(\mathscr{L}_{S,t+1},1_\real)$$
$$=((J^\prime_t\circ(\psi_{H_t},\psi_A))(h_t,a))\circ(\psi_S,\psi_\real)\circ(\mathscr{L}_{S,t+1},1_\real)$$
$$=((J^\prime_t\circ(\psi_{H_t},\psi_A))(h_t,a))\circ(\psi_S\circ\mathscr{L}_{S,t+1},\psi_\real)$$
$$=((J^\prime_t\circ(\psi_{H_t},\psi_A))(h_t,a))(\mathscr{L}_{S^\prime,t+1}\circ\psi_H,1_\real\circ\psi_\real)$$
$$=(((J^\prime_t\circ(\psi_{H_t},\psi_A))(h_t,a))\circ(\mathscr{L}_{S^\prime,t+1},1_\real))\circ(\psi_H,\psi_\real)$$
$$=((J^\prime_t((\psi_{H_t},\psi_A)(h_t,a)))\circ(\mathscr{L}_{S^\prime,t+1},1_\real))\circ(\psi_H,\psi_\real)$$
$$=((J_t^\prime(h^\prime_t,a^\prime))\circ(\mathscr{L}_{S^\prime,t+1},1_\real))\circ(\psi_H,\psi_\real)$$
$$=(T^\prime_t(h^\prime_t,a^\prime))\circ(\psi_H,\psi_\real)$$
$$=(T^\prime_t((\psi_H,\psi_A)(h_t,a)))\circ(\psi_H,\psi_\real)$$
$$=((T^\prime_t\circ(\psi_H,\psi_A))(h_t,a))\circ(\psi_H,\psi_\real)$$
therefore, by definition of morphisms in $\mdpcat$, $(\psi_H,\psi_A,\psi_\real)\in\mdpcat(\mdp,\mdp^\prime)$, which means $\forall\psi,\mathbf{M}(\psi)=(\psi_H,\psi_A,\psi_\real)$ is compatible with the action of $\mathbf{M}$ on $J_t,J_t^\prime$, i.e. $\mathbf{M}$ is a well-defined functor from $\nmdpcat$ to $\mdpcat$.
\end{proof}

\section{Proof of Theorem 1}

\begin{definition}[\textbf{Isomorphism}]
A morphism $f:A\to B$ in a category $\mathfrak{A}$ is an isomorphism if there exists a map $g:B\to A$ in $\mathfrak{A}$ such that $g\circ f=1_A$, $f\circ g=1_B$.
\end{definition}

\begin{definition}[\textbf{Natural Transformation}]
Let $\mathfrak{A}$ and $\mathfrak{B}$ be categories and let $\mathfrak{A}\dar{\mathbf{F}}{\mathbf{G}}\mathfrak{B}$ be functors. A natural transformation $\alpha:\mathbf{F}\to \mathbf{G}$ is a family $\left(\mathbf{F}(A)\xrightarrow{\alpha_A}\mathbf{G}(A)\right)_{A\in\mathfrak{A}}$ of morphisms in $\mathfrak{B}$ such that for every morphism $A\xrightarrow{f}A^\prime$ in $\mathfrak{A}$, the square
\begin{center}
    \begin{tikzcd}
        \mathbf{F}(A) \arrow{r}{
        \mathbf{F}(f)} \arrow{d}{\alpha_A} & \mathbf{F}(A^\prime) \arrow{d}{\alpha_{A^\prime}} \\
        \mathbf{G}(A) \arrow{r}{\mathbf{G}(f)} & \mathbf{G}(A^\prime)
    \end{tikzcd}
\end{center}
commutes. The maps $\alpha_A$ are called the components of $\alpha$.
\end{definition}

\begin{definition}[\textbf{Functor Category}]
Let $\mathfrak{A}$ and $\mathfrak{B}$ be categories. Functor category $\funcat(\mathfrak{A},\mathfrak{B})$ is a category whose objects are all the functors from $\mathfrak{A}$ to $\mathfrak{B}$ and morphisms are all the natural transformations between these functors.
\end{definition}

\begin{definition}[\textbf{Natural Isomorphism}]
Let $\mathfrak{A}$ and $\mathfrak{B}$ be categories. A natural isomorphism between functors $\mathbf{F}$ and $\mathbf{G}$ from $\mathfrak{A}$ to $\mathfrak{B}$ is an isomorphism in category $\funcat(\mathfrak{A},\mathfrak{B})$. Functors $\mathbf{F},\mathbf{G}\in\funcat(\mathfrak{A},\mathfrak{B})$ are isomorphic (denoted as $\mathbf{F}\cong\mathbf{G}$) if there exists a natural isomorphism between them.
\end{definition}

\begin{definition}[\textbf{Category Equivalence}]
A equivalence between categories $\mathfrak{A}$ and $\mathfrak{B}$ consists of a pair of functors together with natural isomorphisms $\eta:1_\mathfrak{A}\to\mathbf{G}\circ\mathbf{F}, ~\epsilon: \mathbf{F}\circ \mathbf{G}\to1_\mathfrak{B}$. If there exists an equivalence between $\mathfrak{A}$ and $\mathfrak{B}$, we say that $\mathfrak{A}$ and $\mathfrak{B}$ are equivalent through $\mathbf{F}$ and $\mathbf{G}$.

\end{definition}

\begin{theorem}[\textbf{Equivalence of $\mdpcat$ and $\nmdpcat$}]
~Category $\mdpcat$ and $\nmdpcat$ are equivalent through functor $\mathbf{M}$ and $\mathbf{N}$, i.e.
$\mathbf{M}\circ\mathbf{N}\cong 1_\mdpcat$,~
$\mathbf{N}\circ\mathbf{M}\cong 1_\nmdpcat$.
\end{theorem}

\begin{proof}


For any
$$\mdp=\mdpdef\in\mdpcat$$
where
$$T_t: (s_t,a_t)\mapsto((s_{t+1},r_t)\mapsto\mathbb{P}_\mdp^t(s_t,a_t; s_{t+1},r_t))$$
we have
$$(\mathbf{M}\circ\mathbf{N})(\mdp)=\mdpdefp$$
where
$$\rho_0^\prime=\rho_0,~S^\prime=H,~A^\prime=A$$
$$T^\prime_t: (h^\prime_t,a_t)\mapsto((h^\prime_{t+1},r_t)\mapsto\mathbb{P}_\mdp^t(s_t,a_t; s_{t+1},r_t))$$
$$h^\prime_t\coloneqq(\{h_\tau\}_{\tau=0}^t,\{a_\tau\}_{\tau=0}^{t-1},\{r_\tau\}_{\tau=0}^{t-1})$$
$$h_t\coloneqq(\{s_\tau\}_{\tau=0}^t,\{a_\tau\}_{\tau=0}^{t-1},\{r_\tau\}_{\tau=0}^{t-1})$$
therefore, we can reconstruct the representation of $T_t$ using $T_t^\prime$ without loss of information and vice versa, 
$$(T_t(s_t,a_t))(s_{t+1},r_t)=(T_t^\prime(\mathscr{L}_{S,t}^{-2}(s_t),a_t))(\mathscr{L}_{S,t+1}^{-2}(s_{t+1}),r_t)$$
$$(T_t^\prime(h^\prime_t,a_t))(h^\prime_{t+1},r_t)=(T_t(\mathscr{L}_{S,t}^2(h^\prime_t),a_t))(\mathscr{L}_{S,t+1}^2(h^\prime_{t+1}),r_t)$$
which means $\mathbf{M}\circ\mathbf{N}\cong 1_\mdpcat$.\\
Similarly,
for any
$$\nmdp\in\nmdpdefj\in\nmdpcat$$
where
$$J_t:(h_t,a_t)\mapsto((s_{t+1},r_t)\mapsto\mathbb{P}^t_\nmdp(h_t,a_t;s_{t+1},r_t))$$
we have 
$$(\mathbf{N}\circ\mathbf{M})(\nmdp)=\nmdpdefjp$$
where
$$\rho^\prime_0=\rho_0,~S^\prime=H,~A^\prime=A$$
$$J^\prime_t:(h^\prime_t,a_t)\mapsto((h_{t+1},r_t)\mapsto\mathbb{P}^t_\nmdp(h_t,a_t;s_{t+1},r_t))$$
$$h^\prime_t\coloneqq(\{h_\tau\}_{\tau=0}^t,\{a_\tau\}_{\tau=0}^{t-1},\{r_\tau\}_{\tau=0}^{t-1})$$
$$h_t\coloneqq(\{s_\tau\}_{\tau=0}^t,\{a_\tau\}_{\tau=0}^{t-1},\{r_\tau\}_{\tau=0}^{t-1})$$
therefore we can reconstruct the representation of $J_t$ using $J_t^\prime$ without loss of information and vice versa,
$$(J_t(h_t,a_t))(s_{t+1},r_t)=(J^\prime_t(\mathscr{L}^{-1}_{S,t}(h_t),a_t))(\mathscr{L}_{S,t+1}^{-1}(s_{t+1}),r_t)$$
$$(J_t^\prime(h^\prime_t,a_t))(h_{t+1},r_t)=(J_t(\mathscr{L}_{S,t}(h^\prime_t),a_t))(\mathscr{L}_{S,t+1}(h_{t+1}),r_t)$$
which means $\mathbf{N}\circ\mathbf{M}\cong 1_\nmdpcat$.
\end{proof}

\section{Proof of Theorem 2}

\begin{theorem}[\textbf{Reversibility of HAS Induced by Group Operator}]
\label{th:revgrp}
The HAS $\mathscr{G}_S$ on MDP $\mdp=\mdpdef$ induced by binary operator $\otimes$ is reversible if $(S,\otimes)$ is a group.
\end{theorem}
\begin{proof}
    Define $\mathscr{G}_S^\ast$ as follows which satisfy the reversibility condition of $\mathscr{G}_S$
    $$\mathscr{G}_{S,t}^\ast(\{\mathscr{G}_{S,\tau}(h_\tau)\}_{\tau=0}^t)\coloneqq(\mathscr{G}_{S,t-1}(h_{t-1}))^{-1}\otimes\mathscr{G}_{S,t}(h_{t})=s_t$$
\end{proof}
\textbf{Note}: Actually, the condition that $(S,\otimes)$ forms a group can be relaxed to $(S^\prime,\otimes)$ forms a group where $S\subseteq S^\prime$.
\section{Proof of Theorem 3}

\begin{theorem}[\textbf{Impact of $\mathbf{G}$ on State Dependency Structure}]
\label{th:gondependency}
    For any MDP $\mdp=\mdpdef$, if it is non-degenerate, then the functor $\mathbf{G}$ induced by some group operator on $S$ ensures that the dependency structure $D_{h^n_t}$ of any history $h^n_t\in H_t^n$ of $\mathbf{G}^n(\mdp)$ satisfies $D_{h^n_t}=[t-n, t]\cap\nat$.
\end{theorem}

\begin{proof}
Based on the closure property of the group operator $\otimes$, we repeatedly apply $\mathbf{G}$ to MDP $\nmdp_0\coloneqq\nmdpdef$ to acquire a series of NMDPs $\{\nmdp_i\}_{i=1}^n\coloneqq\{\langle\rho_0,S,A,\{T_t^i\}_{t=0}^\infty\rangle\}_{i=1}^n$.
$$\mdpcat\ni\nmdp_0\xmapsto{\mathbf{G}}\nmdp_1\xmapsto{\mathbf{G}}\nmdp_2\xmapsto{\mathbf{G}}\cdots\xmapsto{\mathbf{G}}\nmdp_n\in\nmdpcat$$

Consider history $h_t^i\in H_{\nmdp_i}, ~(\forall i\in[0,n]\cap\nat)$ with $\mathscr{E}_S(h_t^i)=s^i_{0:t}$.
By the definition of $\mathbf{G}$ we have
\begin{equation}\label{eq:eq1}\{\mathscr{G}_{S,\tau}(h_\tau^i)\}_{\tau=0}^t=\{s^{i+1}_\tau\}_{\tau=0}^t=\mathscr{E}_S(h_t^{i+1})\end{equation}
and 
\begin{equation}\label{eq:eq2}\left\{(\mathscr{G}_{S,\tau}(h_{\tau}^{i}))^{-1}\otimes\mathscr{G}_{S,\tau+1}(h_{\tau+1}^{i})\right\}_{\tau=0}^{t-1}=\left\{s_{\tau+1}^{i}\right\}_{\tau=0}^{t-1}\end{equation}
substituting Equation \ref{eq:eq1} into Equation \ref{eq:eq2} yields
\begin{equation}\label{eq:staterecur}s_{\tau+1}^{i}=(s_{\tau}^{i+1})^{-1}\otimes s_{\tau+1}^{i+1}\end{equation}
Starting from $s_{t}^0$, repeatedly applying Equation \ref{eq:staterecur} by substituting the left side into the right side yields an expression of $s_{t}^0$ in terms of $s^n_t,s^n_{t-1},\cdots,s^n_{t-n}$:

\newcommand{\s}[2]{s^{#1}_{#2}}
\newcommand{\iv}[1]{(#1)^{-1}}

$$\shr\shr\shr\shr\shr\shr
\begin{array}{rcl}
\s{0}{t} \shr& = &\shr \iv{\s{1}{t-1}}\s{1}{t} \\
\shr& = &\shr \iv{\iv{\s{2}{t-2}}\s{2}{t-1}}(\iv{\s{2}{t-1}}\s{2}{t}) \\
\shr& = &\shr \iv{\s{2}{t-1}}\s{2}{t-2}\iv{\s{2}{t-1}}\s{2}{t} \\
\shr& = &\shr \iv{\iv{\s{3}{t-2}}\s{3}{t-1}}(\iv{\s{3}{t-3}}\s{3}{t-2})\iv{\iv{\s{3}{t-2}}\s{3}{t-1}}(\iv{\s{3}{t-1}}\s{3}{t}) \\
\shr& = &\shr \iv{\s{3}{t-1}}\s{3}{t-2}\iv{\s{3}{t-3}}\s{3}{t-2}\iv{\s{3}{t-1}}\s{3}{t-2}\iv{\s{3}{t-1}}\s{3}{t} \\
\shr& = &\shr \cdots

\end{array}
$$

By mathematical induction, it can be shown that 
\begin{equation}
\s{0}{t}=\bigotimes_{i=1}^{2^n}(\s{n}{t_i})^{2{\chi_{2\mathbb{Z}}}(i)-1}
\end{equation}
where $\{t_i\}_{i=1}^{2^n}\subseteq[t-n,t]\cap\mathbb{Z}$,
$\s{n}{t_i}\coloneqq e, ~(\forall t_i<0)$, $e$ is the unit element of the group.\\
Because $\s{0}{t}$ has the above form, it is sufficient to specify the group operator as the multiplication in the free group to obtain the theorem.
\end{proof}

\textbf{Note}: Free group is not the only form in which $(S,\otimes)$ can be. For example, the theorem still holds when the group is $(\real^k,+), ~\forall k\in\natplus$.

\section{Proof of Theorem 4}
\begin{theorem}[\textbf{Reversibility of HAS induced by Auxiliary Sequence and Convolution Operator}]
The HAS $\mathscr{R}_S\coloneqq\{\mathscr{R}_{S,t}\}_{t=0}^\infty$ induced by prefixes of auxiliary sequence $\{w_t\}_{t=0}^\infty$ and operator $\laast$ is reversible if $w_0$ is invertible in the ring.
\end{theorem}
\begin{proof}
Define $\mathscr{R}^\ast_S$ as follows which satisfy the reversibility condition of $\mathscr{R}_S$ if $w_t$ is inversible.
$$
\shr\mathscr{R}_{S,t}^\ast(\{\mathscr{R}_{S,\tau}(h_\tau)\}_{\tau=0}^t)\coloneqq w_t^{-1}\cdot(\mathscr{R}_{S,t}(h_t)\oplus(-\mathscr{R}_{S,t-1}(h_{t-1})))=s_t
$$
\end{proof}

\section{Proof of Theorem 5}

\begin{theorem}[\textbf{Reversibility of HAS induced by Auxiliary Sequence and Convolution Operator}]
\label{th:invhasconv}
The HAS $\mathscr{R}_S\coloneqq\{\mathscr{R}_{S,t}\}_{t=0}^\infty$ induced by prefixes of auxiliary sequence $\{w_t\}_{t=0}^\infty$ and operator $\laast$ is reversible if $w_0$ is invertible in the ring.
\end{theorem}

\begin{proof}
By the definition of convolution operator $\laast$, we have
$$\mathscr{R}_{S,\tau}(h_\tau)=\bigoplus_{i=0}^\tau w_i\cdot s_{\tau-i}$$
the following equation comes from taking $\tau\in[0,t]\cap\nat$
\begin{equation}\shr\label{eq:convlin}\shr\begin{array}{rcl}\left[\begin{matrix}
    w_0 & w_1 & w_2 & \cdots & w_t     \\
    0   & w_0 & w_1 & \cdots & w_{t-1} \\
    0   &   0 & w_0 & \cdots & w_{t-2} \\
    \cdots   &   \cdots &  \cdots  & \cdots & \cdots  \\
    0   &   0 &   0 & \cdots & w_0
\end{matrix}\right]\cdot\left[\begin{matrix}
s_t \\ s_{t-1} \\ s_{t-2} \\ \cdots \\ s_0
\end{matrix}\right] & = & \left[\begin{matrix}
\mathscr{R}_{S,t}(h_t) \\ \mathscr{R}_{S,t-1}(h_{t-1}) \\ \mathscr{R}_{S,t-2}(h_{t-2}) \\ \cdots \\ \mathscr{R}_{S,0}(h_0)
\end{matrix}\right]\\
\mathbf{w}\cdot \mathbf{s} & = & \mathbf{r}
\end{array}
\end{equation}
If $w_0$ is invertible in the ring, the Gaussian elimination method can be used to solve Equation \ref{eq:convlin} and obtain the expression for $s_t$, which also means the matrix $\mathbf{w}$ is invertible in the ring.
\end{proof}

\section{Proof of Theorem 6}
\begin{theorem}[\textbf{Impact of $\mathbf{R}$ on State Dependency Structure}]
\label{th:rondependency}
For any MDP $\mdp=\mdpdef$, if it is non-degenerate, then the functor $\mathbf{R}$ induced by prefixes of auxiliary sequence $\{w_\tau\}_{\tau=0}^\infty$ and operator $\laast$ ensures that the dependency structure $D_{h^\prime_t}$ of any history $h^\prime_t\in H^\prime_t$ of $\mathbf{R}(\mdp)$ satisfies $D_{h^\prime_t}=\{t-\tau\mid(\mathbf{w}^{-1})_{0,\tau}\neq0\}$, where $\mathbf{w}^{-1}$ is the inverse matrix of $\mathbf{w}$ in the ring, $0$ is the zero element of the ring.
\end{theorem}

\begin{proof}
As is proved in Theorem \ref{th:invhasconv}, $\mathbf{s}=\mathbf{w}^{-1}\mathbf{r}$, where
\begin{equation}\shr
\mathbf{w}^{-1}=
\left[\begin{matrix}
    w_0^{-1} & -w_0^{-1}\cdot w_1\cdot w_0^{-1} & \cdots & \cdots & \cdots     \\
    0 & w_0^{-1} & \cdots & \cdots & \cdots \\
    \cdots & \cdots & \cdots & \cdots & \cdots \\
    0 & \cdots & \cdots & w_0^{-1} & -w_0^{-1}\cdot w_1\cdot w_0^{-1}  \\
    0 & \cdots & \cdots & 0 & w_0^{-1}
\end{matrix}\right]
\end{equation}
is an upper-triangular matrix,
therefore
\begin{equation}
    \label{eq:convst}
    s_t=\bigoplus_{\tau=0}^t (\mathbf{w}^{-1})_{0,\tau}\cdot\mathscr{R}_{S,t-\tau}(h_{t-\tau})
\end{equation}

If $(\mathbf{w}^{-1})_{0,\tau}$ is not zero, then there exists some aggregated state $s^\prime_{t-\tau}\in\mathscr{R}(H)$ that will change $s_t$ when using it to replace $\mathscr{R}_{S,t-\tau}(h_{t-\tau})$ in Equation \ref{eq:convst}, which means $t-\tau\in D_{h_t^\prime}$.
\end{proof}

\section{History Aggregator for Reward (HAR)}
Unlike HAS, which introduces non-Markovianity into the MDP's transition dynamics, HAR does so in the MDP's reward mechanisms.
HAR is similar to reward shaping. However, instead of simplifying RL training as reward shaping does, it is used to construct NMDPs with non-Markovian reward mechanisms.

\begin{definition}[\textbf{History Aggregator for Reward (HAR)}]
For any MDP $\mdp=\mdpdef$, an HAR $\ha{\real}{}\coloneqq\hadef{\real}{}$ is a series of maps in which $\mathscr{A}_{\real,t}:H_{t+1}\to\real,~(\forall t\in\nat)$.
\end{definition}
\begin{definition}[\textbf{Reversibility of HAR on MDP}]
 An HAR $\ha{\real}{}=\hadef{\real}{}$ on MDP $\mdp=\mdpdef$ is reversible iff there exists a series of maps denoted as $\ha{\real}{\ast}\coloneqq\hadef{\real}{\ast}$ which satisfy:
$$\mathscr{A}_{\real,t}^\ast(\{\mathscr{A}_{\real,\tau}(h_{\tau+1})\}_{\tau=0}^t)=r_t,~~(\forall t\in\nat)$$
where $h_\tau\in H_\tau, ~h_1\prec h_2\prec\cdots\prec h_{t+1}, ~r_t=\mathscr{L}_{\real,t}(h_{t+1})$.
\end{definition}

\begin{definition}[\textbf{Application of Reversible HAR to MDP}]
The application of a reversible HAR $\ha{\real}{}=\hadef{\real}{}$ to an MDP $\mdp=\mdpdef$ is an NMDP $\nmdp^\prime=\nmdpdefp$, $\rho_0^\prime=\rho,S^\prime=S,A^\prime=A$ in which:
 $$T_t^\prime: \begin{array}{rcl}\shr H_t^\prime\times A^\prime & \to & \Delta_{S^\prime\times\real} \\
    \shr(h_t^\prime,a) & \mapsto & ((T_t\circ(\mathscr{L}_{S^\prime,t}, 1_A))(h^\prime_t,a))\circ\left(1_{S^\prime}, G_{\real, h_t^\prime}\right)
    \end{array}$$
    where $G_{\real,h_t^\prime}: \begin{array}{rcl}
    \real & \to & \real \\
    r_t^\prime & \mapsto & \ha{\real,t}{\ast}\left(\mathscr{E}_{\real}(h^\prime_t), r_t^\prime\right)
    \end{array}$.
\end{definition}
\begin{corollary}
The series of latest reward extraction operators, $\mathscr{L}_\real\coloneqq\{\mathscr{L}_{\real,t}\}_{t=0}^\infty$, constitutes a reversible HAR on MDP.
\end{corollary}

\section{Extending Reversible HAS to be Applicable on NMDP}

\begin{itemize}
\item \textbf{HAS}: 
Although HAS was previously described as a structure established on MDP, due to the formal similarity between MDP and NMDP, we can directly replace all occurrences of "MDP" in this definition with "MDP or NMDP".
This still constitutes a valid definition.\newline
\item \textbf{Reversibility of HAS}: 
Unlike HAS on MDP, to ensure that the target NMDP can utilize the transition function from the original NMDP, it must be possible to reconstruct the complete state sequence from the aggregated history sequence, rather than just the current state.
Therefore, a reverse $\ha{S}{\ast}=\hadef{S}{\ast}$ of an HAS $\ha{S}{}=\hadef{S}{}$ on NMDP $\nmdp=\nmdpdef$ should satisfy:
$$\shr\!\!\ha{S,t}{\ast}(\{\ha{S,\tau}{}(h_\tau)\}_{\tau=0}^\infty)=\hat{h}_t\in\mathscr{E}_S^{-1}(\mathscr{E}_S(h_t))\subseteq H_t, ~(\forall t\in\nat)$$
\item \textbf{Application of Reversible HAS}:
The application of reversible HAS on NMDP is similar to that on MDP, with $G_{S,h^\prime_t}$ replaced by the following:
$$G_{S,h^\prime_t}:\begin{array}{rcl}
    S^\prime & \to & S\\
    s_{t+1}^\prime & \mapsto & \left(\mathscr{L}_{S,t+1}\circ\ha{S,t+1}{\ast}\right)\left(\mathscr{E}_{S^\prime}(h_t^\prime),s^\prime_{t+1}\right)
\end{array}$$
The use of $\mathscr{L}_{S,t+1}$ is to extract the latest state from the history returned by $\ha{S,t+1}{\ast}$, according to the definition of reversed HAS on NMDP.\newline
\item \textbf{HAS Induced by Binary Operator}:
Similar to extending the concept of HAS from MDP to NMDP, in this definition, "MDP" can also be replaced with "MDP or NMDP", which still constitutes a valid definition.\newline
\end{itemize}

\section{Intuitive Explanation and Examples of Key Concepts}
\subsection{MDP and NMDP}
\begin{figure}[H]
\includegraphics[width=1.1\columnwidth]{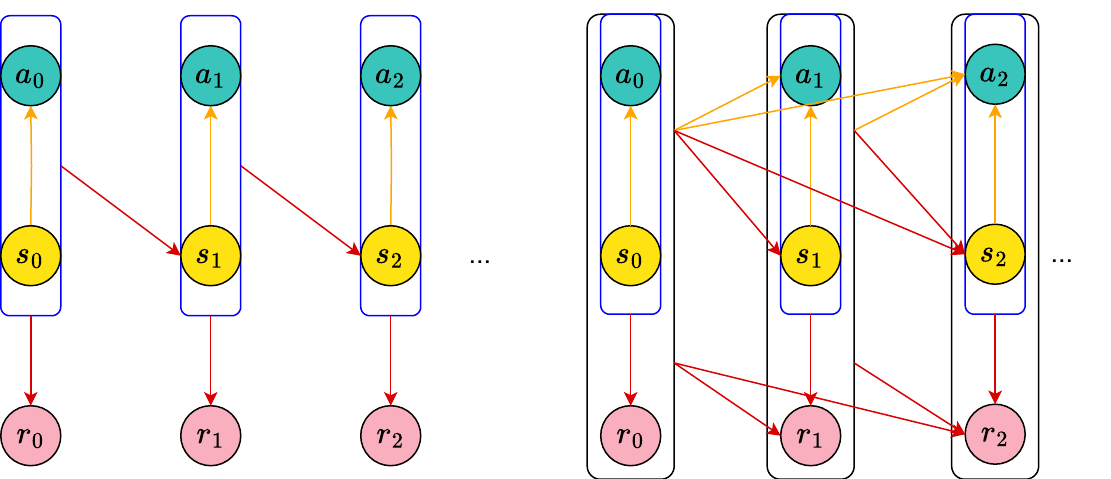}
\caption{
Probabilistic graphical model of policy and transitions in a standard MDP (left) and an NMDP (right).
red arrows indicate the transition dynamics of the environment, while orange arrows represent the policy. Arrows from a frame indicate probabilistic dependencies between all the framed variables and the variable at the head of the arrow.}
\label{time_seq}
\end{figure}

Figure.\ref{time_seq} illustrates the probabilistic dependencies between states, actions, and rewards in the definitions of MDP and NMDP. Note that, by definition, the transition dynamics of MDPs and NMDPs return a joint distribution over the next state and reward. For clarity, Figure \ref{time_seq} simplifies this representation to emphasize the main conditional independence relationships.

As illustrated, the key difference between MDP and NMDP lies in the factors determining state transitions and rewards. In an MDP, both the transition to the next state and the received reward rely solely on the current state. For example, in a simple robotic movement model, the robot's next position and the reward it gets are only determined by its current location.
Conversely, in an NMDP, the transition to the next state and the associated reward are influenced by previous states, actions, and rewards. This means that historical information plays a crucial role in the decision-making process.

A typical example of an NMDP is stock price prediction in the financial market. Consider a particular stock. The "state" here can be defined by various factors such as the current stock price, trading volume, and relevant market indices. The "action" could be decisions made by investors, like buying, selling, or holding the stock.
The future stock price (the next state) is not simply determined by the current stock price (the current state) and the current investment actions (the current action). Instead, it is highly correlated with historical data. For instance, if a stock has shown a consistent upward trend over the past few quarters due to strong company fundamentals and positive market sentiment, this historical pattern is a significant factor in predicting the stock price in the next period. Also, past events such as quarterly earnings announcements, major product launches, or changes in industry regulations can all leave an imprint on the stock's price movement history and have a substantial impact on future price predictions.

\subsection{The Category-Theoretic Outlook on MDP and NMDP}
\begin{itemize}

\item A \textbf{category} is composed of a collection of objects (items) and a collection of morphisms (also known as maps or arrows). Each morphism has a domain and a codomain, that is, each arrow has a source and a target.
\\
\item \textbf{Object}s in a category can be anything. For example, all sets form the category of sets, all groups form the category of groups, and all topological spaces form the category of topological spaces.
\\
\item \textbf{Morphism}s represent the connections between pairs of objects. These connections take different forms in different categories. In the category of sets, it is a function; in the category of partially ordered sets, it is a partial order relation; in a category constructed from a set, it is an equivalence relation; in the category of groups, it is a homomorphism; in the category of linear spaces, it is a linear map; or it can be any connection that conforms to the essential characteristics of the objects.

In the MDP category $\mdpcat$, a morphism $\phi=\morphism{\phi}$ from $\mdp$ to $\mdp^\prime$ bears the following intuitive meaning:
Map the states, actions, and rewards of $\mdp$ separately using $\morphism{\phi}$.
Thus,  we obtain the MDP $\mdp^\prime$, whose states, actions, and rewards are images under $\phi$ and the transition dynamics (along with the reward mechanics) are identical to those of $\mdp$, except for the different representations of states, rewards, and actions. The intuitive explanation of morphisms in the NMDP category $\nmdpcat$ is similar to that in the MDP category $\mdpcat$.
\\
\item \textbf{Functor}s, on the other hand, represent the connections between pairs of categories. A functor preserves the structure of each category, mapping objects to objects and morphisms to morphisms while maintaining the structure unchanged.
For example, the fundamental group functor maps a topological space to a group. Since it preserves the structure of the topological space, the study of the topological space using geometric methods is transformed into the study of the fundamental group using algebraic methods.
\\
\item Both \textbf{category isomorphism} and \textbf{category equivalence} capture a sense of "similarity" between categories, but they diverge in key aspects.
Category isomorphism represents a special and highly restrictive form of categorical equivalence. Isomorphic categories are, by default, equivalent. They both signify a profound structural and property-based equivalence between categories.
Category equivalence, on the other hand, offers more flexibility. It amounts to an "approximate one-to-one correspondence" and is realized through the interaction of two functors. While it preserves many crucial properties, it does so less rigidly than category isomorphism.
Take, for example, the category of finite-dimensional vector spaces and the category of matrices. Finite-dimensional vector spaces can be related to matrices, with linear transformations corresponding to matrix multiplications. Although this is not a one-to-one mapping, the two categories are equivalent.

The equivalence between MDP category $\mdpcat$ and NMDP category $\nmdpcat$   implies the following: For any MDP, the non-Markov embedding functor $\mathbf{N}$ designates an NMDP.
In the NMDP category, this NMDP can maintain the structure created by the morphisms related to the original MDP in the MDP category. 
Conversely, for any NMDP, the Markov abstraction functor $\mathbf{M}$  determines an MDP.
In the MDP category, this MDP can preserve the structure formed by the morphisms associated with the original NMDP in the NMDP category.
Notably, the non-Markov embedding functor $\mathbf{N}$ is not the sole means of choosing an NMDP for an MDP, and the Markov abstraction functor $\mathbf{M}$  isn't the only way to select an MDP for an NMDP either.
\end{itemize}
\subsection{The Reversibility Condition of HAS}
As the name implies, the History aggregator for State (HAS) constructs the NMDP state by aggregating the MDP history.
The reversibility condition of HAS ensures that a decision-making algorithm can reconstruct the original MDP state from the history of the NMDP.
\begin{example}
We have a non-degenerate MDP $\mdp=\mdpdef$, where $S=\real^n$, and a reversible HAS $\mathscr{S}_{S,t}: h_t\mapsto\sum_{\tau=0}^t s_\tau$ together with its reverse: $$\begin{array}{rcl}\mathscr{S}_{S,t}^\ast(\{\mathscr{S}_{S,\tau}(h_\tau)\}_{\tau=0}^t) & \coloneqq & \mathscr{S}_{S,t}(h_t)-\mathscr{S}_{S,t-1}(h_{t-1}) \\ & = & s^\prime_t-s^\prime_{t-1}\\& = & \sum_{\tau=0}^t s_\tau-\sum_{\tau=0}^{t-1} s_\tau\\ & = & s_t\end{array}$$
which recover the MDP state $s_t$ from the NMDP state series $s^\prime_{0:t}=\{\mathscr{S}_{S,\tau}(h_\tau)\}_{\tau=0}^t$.\\
Applying $\mathscr{S}_{S,t}$ to $\mdp$, we get an NMDP $\nmdp=\nmdpdefp$, where 
$\rho_0^\prime=\rho_0,~S^\prime=S,~A^\prime=A$,
$$(T_t^\prime(h_t^\prime,a_t))(s^\prime_{t+1},r_t)=(T_t(s_t,a_t))(s_{t+1},r_t)$$
For any history $h_t=(s_{0:t},a_{0:t-1},r_{0:t-1})$ of $\mdp$, the correspond history of $\nmdp$ is $$\begin{array}{rcl}h_t^\prime & = & (\{\mathscr{S}_{S,\tau}(h_\tau)\}_{\tau=0}^t,a_{0:t-1},r_{0:t-1}) \\ & = & (s^\prime_{0:t},a_{0:t-1},r_{0:t-1})\\ & = & (\{\sum_{i=0}^\tau s_i\}_{\tau=0}^t,a_{0:t-1},r_{0:t-1})\end{array}$$
The state dependency structure of $h^\prime_t$ is $\{t, t-1\}$ because calculating $T^\prime_t(h^\prime_t,a_t)$ which is equivalent to calculating $T_t(s_t,a_t)$ or $s_t$ only requires $s^\prime_t=\sum_{\tau=0}^t s_\tau$ and $s^\prime_{t-1}=\sum_{\tau=0}^{t-1}s_\tau$.
\end{example}

The reversibility condition is essential for reconstructing an MDP's state from an NMDP's history. When this condition is violated, consider the use of the non-reversible HAS $\mathscr{B}_{S,t}:h_t\mapsto s_t\odot e_1$, where $e_1=(1,0,...,0)\in\real^n$, and the operator $\odot$ is defined as 
$$\odot:~\begin{array}{ccc}\real^n\times\real^n\ & \to & \real^n\\((x_1,...x_n), (y_1,..,y_n)) & \mapsto & (x_1y_1,...,x_ny_n)\end{array}$$
In such a scenario, the non-reversible HAS retains only the first-dimension value of $s_t$ and discards all others. For any decision-making algorithm designed to solve the NMDP, it is impossible to reconstruct the MDP state $s_t$ using the NMDP history $h^\prime_t$.
Although the functor induced by $\mathscr{B}_S$ can be easily implemented as an environment wrapper, it transforms an MDP into a partially-observable NMDP instead of an NMDP.
The absence of necessary information thus increases the complexity of the original MDP.

\subsection{Reversible HAS Induced by Group Operator}
The intuition for the reversible HAS induced by a group operator is derived from the concept of the prefix-sum of a sequence. It can be shown that the prefix-sum of a time-homogeneous first-order Markov chain is a time-homogeneous second-order Markov chain.
\begin{theorem}
Let $\{X_n\}_{n=0}^\infty$ be a time-homogeneous first-order Markov chain, then $\{S_n\}_{n=0}^\infty\coloneqq\{\sum_{i=0}^n X_i\}_{n=0}^\infty$ is a time-homogeneous second-order Markov chain.
\end{theorem}
\begin{proof}
Since $\{X_n\}_{n=0}^\infty$ is a time-homogeneous first-order Markov chain, we have
$$\begin{array}{l}\mathbb{P}(X_{n+1}=x_{n+1}\mid X_n=x_n,...,X_1=x_1) \\ =\mathbb{P}(X_{n+1}=x_{n+1}\mid X_n=x_n)\end{array}$$
Then $$\begin{array}{l}\mathbb{P}(S_{n+1}=s_{n+1}\mid S_n=s_n,...,S_1=s_1) \\ = \mathbb{P}(S_{n+1}-S_n=s_{n+1}-s_n\mid S_n=s_n,...,S_1=s_1) \\ = \mathbb{P}(X_{n+1}=s_{n+1}-s_n\mid S_n=s_n,...,S_1=s_1) \\ = \mathbb{P}(X_{n+1}=s_{n+1}-s_n\mid X_n=s_n-s_{n-1},...,X_1=s_1) \\ = \mathbb{P}(X_{n+1}=s_{n+1}-s_n\mid X_n=s_n-s_{n-1})\end{array}$$
And $$\begin{array}{l}\mathbb{P}(S_{n+1}=s_{n+1}\mid S_n=s_n,S_{n-1}=s_{n-1}) \\ = \mathbb{P}(S_{n+1}-S_n=s_{n+1}-s_n\mid S_n=s_n,S_{n-1}=s_{n-1}) \\ = \mathbb{P}(X_{n+1}=s_{n+1}-s_n\mid X_n=s_n-s_{n-1},S_{n-1}=s_{n-1}) \\ = \mathbb{P}(X_{n+1}=s_{n+1}-s_n\mid \{X_n=s_n-s_{n-1}\}\cap\{ \\ \sum_{i=1}^{n-1}X_i=s_{n-1}\}) \\ = \mathbb{P}(X_{n+1}=s_{n+1}-s_n\mid \{X_n=s_n-s_{n-1}\}\cap \\ \bigsqcup_{\sum_{i=1}^{n-1}x_i=s_{n-1}}\{X_{n-1}=x_{n-1},...,X_1=x_1\}) \\ = \mathbb{P}(X_{n+1}=s_{n+1}-s_n\mid \bigsqcup_{\sum_{i=1}^{n-1}x_i=s_{n-1}}\{\\X_n=s_n-s_{n-1},X_{n-1}=x_{n-1},...,X_1=x_1\}) \\ = \mathbb{P}(X_{n+1}=s_{n+1}-s_n\mid X_n=s_n-s_{n-1})\end{array}$$
Therefore we have $$\begin{array}{l}\mathbb{P}(S_{n+1}=s_{n+1}\mid S_n=s_n,...,S_1=s_1) \\ = \mathbb{P}(S_{n+1}=s_{n+1}\mid S_n=s_n,S_{n-1}=s_{n-1})\end{array}$$
which means $\{S_n\}_{n=0}^\infty$ is a time-homogeneous second-order Markov chain.
\end{proof}

Similarly, it can be shown that the prefix-sum of a time-homogeneous $n$-th order Markov chain is a time-homogeneous $n+1$-th order Markov chain.
By repeatedly applying the prefix-sum operation to a time-homogeneous first-order Markov chain, we can obtain a series of Markov chains with increasing orders.
We generalize the prefix-sum operation to aggregation by using the group operator, with MDPs/NMDPs instead of Markov chains as operands. This leads to the development of the reversible HAS induced by the group operator.

\subsection{Reversible HAS Induced by Auxiliary Sequence and Convolution Operator}
The convolution operation in signal processing serves as the source of inspiration for reversible HAS induced by auxiliary sequence and convolution operator. The convolution of two sequence $\{x_n\}_{n=0}^\infty$ and $\{y_n\}_{n=0}^\infty$ is $$\{x_n\}_{n=0}^\infty\ast\{y_n\}_{n=0}^\infty=\{\sum_{i=0}^n x_ny_{n-i}\}_{n=0}^\infty$$

In the above equation, we fix $\{x_n\}_{n=0}^\infty$ as a constant sequence, let $\{y_n\}_{n=0}^\infty$ be the sequence of states in the history of the MDP, and define multiplication and addition appropriately, thus forming the expression form of HAS induced by auxiliary sequence and convolution operator.
\section{The General Applicability of HAS}
\subsection{Reversible HAS Induced by Group Operator}
As is proved in Theorem \ref{th:gondependency}, if the group is a free group or $(\real^k,+)$ for some $k$, then applying functor $\mathbf{G}$ $n$ times ensures that the state dependency structure is $[t-n,t]\cap\nat$, which means for all $i\in [t-n,t]\cap\nat$, the NMDP state $s^\prime_i$ is essential for the decoding of MDP state $s_t$ from the NMDP history.

Although there are multiple ways to extend $S$ into a group, as can be seen from the proof of Theorem \ref{th:gondependency}, not every approach can ensure that all elements in the above-mentioned state dependency structure $[t-n,t]\cap\nat$ play a role in the decoding process. Therefore, using other extension methods may result in the actually obtained state dependency structure being “sparser” compared to the case of extending $S$ to $(\real^k,+)$ for some $k$.

The reversible HAS induced by group operator can only create state dependency structure in the form of $[t-n,t]\cap\nat$, and it fails to provide a method for specifying the “degree” of each item's dependency.
This is also a drawback of this approach.
\subsection{Reversible HAS Induced by Auxiliary Sequence and Convolution Operator}
As proven in Theorem \ref{th:rondependency}, in this case, the state dependency structure is given by $\{t-\tau\mid(\mathbf{w}^{-1})_{0,\tau}\neq0\}$, and the "degree" of dependency of $t-\tau$ is $(\mathbf{w}^{-1})_{0,\tau}$.
Thus, we can assign any zero or non-zero, large or small values to the entries of the upper-triangular matrix $\mathbf{w}^{-1}$ to control the dependency structure and the "degree" of dependency, making it a flexible way of constructing dependency structure into MDPs.

However, the HAS induced by the auxiliary sequence and convolution operator treats all MDP histories of the same length equally, which means that it assigns the same weights to histories of the same length.

\subsection{Further Generalization}
Within the framework of reversible HAS, the constructions using group operators and auxiliary sequences are merely two conveniently implementable special cases.
There are clear theoretical guarantees for their properties and well-defined methods to analyze their dependency structures.
However, they are by no means the only way of constructing reversible HAS.
We can also consider non-linear sequence processing methods and sequence processing methods that are sensitive to different histories, etc.

When the reversibility condition is abandoned, HAS becomes a general theoretical model for constructing general partially observable NMDPs. We can use methods such as information hiding and introducing randomness to ensure that the resulting partially observable NMDPs possess the desired properties.

In summary, HAS offers a general theoretical model for transforming MDPs into NMDPs. The reversibility condition captures the key to ensuring the essential equivalence between the pre-transformation and post-transformation MDPs. The HAS framework still holds vast exploration potential and promising research prospects in future studies.